\UseRawInputEncoding
\documentclass[acmsmall]{acmart}

\usepackage{amsthm}
\usepackage{graphicx}
\usepackage[rightcaption]{sidecap}
\usepackage{wrapfig}
\usepackage{multirow}
\usepackage{makecell}
\usepackage{float}
\usepackage{threeparttable}

\AtBeginDocument{%
  \providecommand\BibTeX{{%
    \normalfont B\kern-0.5em{\scshape i\kern-0.25em b}\kern-0.8em\TeX}}}

\setcopyright{acmcopyright}
\copyrightyear{2021}
\acmYear{2021}

\acmJournal{JACM}

\begin{document}

\title{What Can Knowledge Bring to Machine Learning? ¨C- A Survey of Low-shot Learning for Structured Data}

\author{Yang Hu}
\email{Yang.Hu@soton.ac.uk; superhy199148@hotmail.com}
\authornote{Corresponding author.}
\orcid{0000-0002-4856-5014}
\affiliation{%
  \institution{University of Southampton}
  \country{United Kingdom}
}
\affiliation{%
  \institution{South China University of Technology}
  \streetaddress{Building B-3, University City}
  \city{Guangzhou}
  \state{Guangdong}
  \postcode{510006}
  \country{China}
}

\author{Adriane Chapman}
\email{Adriane.Chapman@soton.ac.uk}
\affiliation{%
  \institution{University of Southampton}
  \streetaddress{University Road}
  \city{Southampton}
  \state{Hampshire}
  \postcode{SO17 1BJ}
  \country{United Kingdom}
}

\author{Guihua Wen}
\email{crghwen@scut.edu.cn}
\affiliation{%
  \institution{South China University of Technology}
  \streetaddress{Building B-3, University City}
  \city{Guangzhou}
  \state{Guangdong}
  \postcode{510006}
  \country{China}
}

\author{Dame Wendy Hall}
\email{wh@ecs.soton.ac.uk}
\affiliation{%
  \institution{University of Southampton}
  \streetaddress{University Road}
  \city{Southampton}
  \state{Hampshire}
  \postcode{SO17 1BJ}
  \country{United Kingdom}
}

\renewcommand{\shortauthors}{Hu, Chapman, Hall}

\begin{abstract}
 Supervised machine learning has several drawbacks that make it difficult to use in many situations. Drawbacks include: heavy reliance on massive training data, limited generalizability and poor expressiveness of high-level semantics. Low-shot Learning attempts to address these drawbacks. Low-shot learning allows the model to obtain good predictive power with very little or no training data, where structured knowledge plays a key role as a high-level semantic representation of human. This article will review the fundamental factors of low-shot learning technologies, with a focus on the operation of structured knowledge under different low-shot conditions. We also introduce other techniques relevant to low-shot learning. Finally, we point out the limitations of low-shot learning, the prospects and gaps of industrial applications, and future research directions.
\end{abstract}


\maketitle

\section{Introduction}
\label{sec1}

Throughout the past 30 years of machine learning (ML), various artificial intelligence (AI) models were created to better utilize large amounts of training data.  ML models based on massive data training have created a benchmark for the perception of the art of the possible. Vivid examples include the ImageNet competition~\cite{russakovsky2015imagenet} and its champions that perform better than humans~\cite{szegedy2015going,he2016deep,hu2018squeeze}, language models obtained through massive text  training~\cite{mikolov2013efficient,pennington2014glove,devlin2019bert,yang2019xlnet}, and various speech data sets~\cite{engel2017neural,gemmeke2017audio,ardila2019common}.  AlphaGo~\cite{silver2016mastering} has defeated the human Go champion, and its upgraded version of AlphaGo Zero~\cite{silver2017mastering}, which is entirely free from chess training.

Despite the fantastic results obtained by the above models on specific data sets, usage of these techniques still face  challenges including: 1. Large-scale data collection and labeling consume a lot of labor; Related to gathering data, it is also challenging to find a closed game rule space like Go.  2. The generalization ability of the traditional ML model is limited. For many models trained on specific data sets, some simple attacks~\cite{goodfellow2014explaining,thys2019fooling} can easily crash their original excellent performance, which indicates the gap between reality to general AI.

In response to above issues, an intuitive idea is to develop techniques that use tiny or even zero-scale training samples for learning. Humans usually only need very few or even zero samples to learn new concepts~\cite{li2002rapid,thorpe1996speed}, which is lacking in traditional ML model. In this review, we call such technique Low-shot Learning (LSL), it should be noted that Low-shot Learning actually contains two directions, the first one is few-shot learning (FSL)~\cite{wang2020generalizing,shu2018small}, which requires little or even only one training sample to complete the training of the ML model. FSL includes as a special case one-shot learning (OSL)~\cite{fei2006one}). In this research direction, researchers mainly apply the meta-learning (Meta-L) strategy~\cite{lemke2015metalearning} to endow the new model the ability to optimize rapidly with the existing training knowledge. These methods suggest the idea of "learning to learn"~\cite{bengio1991learning}. Another direction, zero-shot learning (ZSL)~\cite{wang2019survey}, requires the model to predict samples that have never seen before. Since it is impossible to obtain any training data of predicted objects, the ZSL model establishs the linkages between unseen objects and other seen objects to realize the transfer of semantic knowledge~\cite{lampert2013attribute}. To sum up, the above two directions of LSL represent the trend of  ML from the data-driven perception era to the cognitive age of high-level semantic representation. LSL help AI systems get rid of the dependence on massive data and attempt to create more generalizable solutions by learning relationships.

Although low-shot learning was first used in image recognition tasks and achieved rapid development in the field of computer vision (CV), it also has good prospects in the field of structured data processing such as knowledge extraction and ontology learning~\cite{hazman2011survey,maedche2001ontology,cerbah2008learning,alani2003automatic}, especially in this new era of Internet development~\cite{o2018four}. For example, in the Defense and Security domain, it has been used for cyber defense and security~\cite{colbaugh2011proactive,kou2004survey}, and fraud detection~\cite{buczak2015survey}. The emergence of LSL also brings more possibilities for structured data modeling such as relationship extraction and classification~\cite{yuan2017one,xie2020heterogeneous},  completion of a knowledge base~\cite{zhang2019few}, or  embedding representations of ontology~\cite{zhou2018deep,rios2018few}.

At present, there are already reviews of general FSL~\cite{fu2018recent,wang2020generalizing} and ZSL~\cite{wang2019survey} methods that summarize the two directions of LSL. However, the community still lacks a survey of LSL for structured data modeling. This survey makes the following contributions:

\begin{itemize}
\item We summarize both two branches of low-shot learning -- few-shot learning and zero-shot learning, define their main problem and basic concepts, then connect them through their modeling of knowledge and other structured data.
\item We focus on applying LSL to structured data, including the main goals, approaches, data sets, applications, and challenges. We will not only review the application of LSL in structured data learning but also analyze the role of structured data modeling in promoting LSL.
\item We discuss other technologies relevant to LSL to provide a more in-depth analysis and understanding of LSL from the view of these related technologies.
\item We analyze the gap from state-of-the-art LSL technology in the field of structured data learning to large-scale industrial applications. And from the view of the practical issue, analyze the development trend of LSL and the future impact on ML of structured data.
\end{itemize}

The remaining sections of this review will be organized as follows. In section~\ref{sec2}, we present problem definitions, basic concepts, necessary materials, etc. of LSL, from two branches of FSL and ZSL.  We review the classic and state-of-the-art models in LSL in section~\ref{sec3}. Then we introduce the researches of LSL in the field of structured data learning. Further, we also discuss the role of structured data such as knowledge graph (KG) in LSL tasks of
various research fields. In section~\ref{sec4}, we introduce the datasets designed for, or applied in, LSL tasks. We introduce other technologies relevant to LSL, and we analyze the technical relationship between them and LSL, as well as the similar motivations and design concepts implicitly behind them  in section~\ref{sec5}. Section~\ref{sec6} provides the discussion about the gap from state-of-the-art LSL technologies to the large scale industrial application, then we give the prediction of the development directions of LSL, especially for structured data learning. Section~\ref{sec7} concludes this review.

\section{Overview of Low-shot Learning}
\label{sec2}

\begin{figure}[t]
\centering
\includegraphics[scale=0.45]{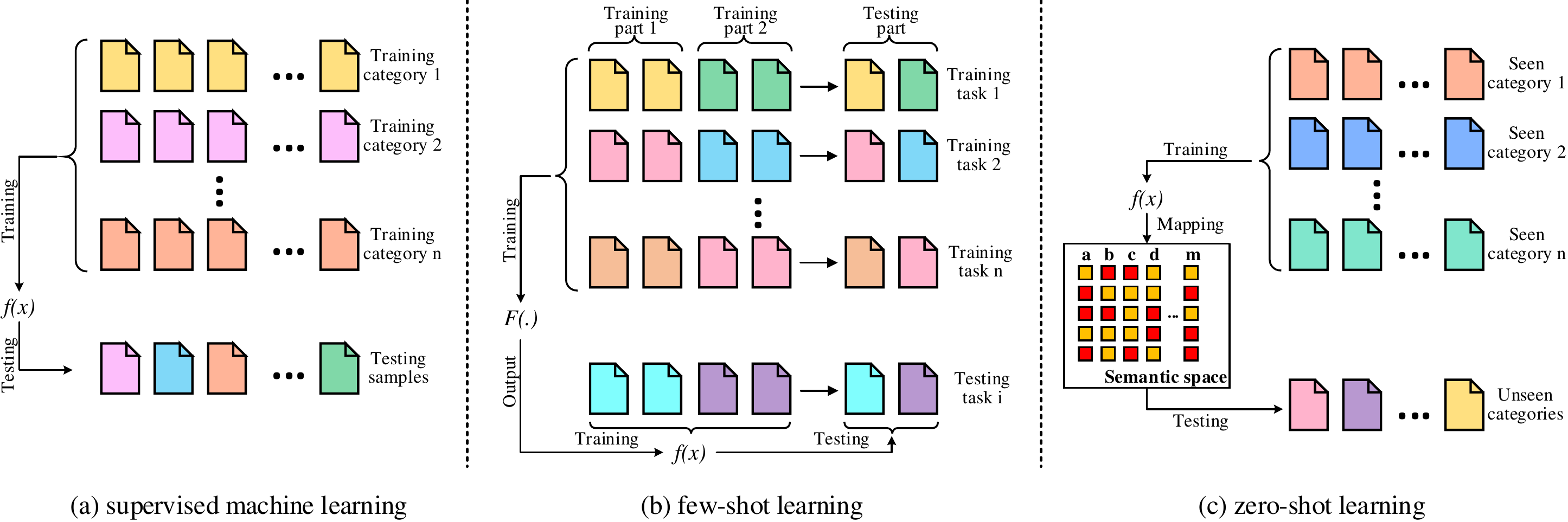}
\caption{Comparison illustration of typical supervised machine learning, few-shot learning, and zero-shot learning. Different colored boxes represent samples from different categories. (a) supervised machine learning with sufficient training data, training and testing data come from the same category space; (b) few-shot learning (using 2-way 2-shot scenario as the example), $F(\cdot)$ is trained to generate a swiftly learning model $f(x)$; (c) zero-shot learning, where $f(x)$ is trained to running prediction through the semantic space.}
\label{fig2-1}
\end{figure}

LSL is the exploration of the future development direction of machine learning, especially supervised learning. As the special branch of machine learning, it can be widely used in various research and application fields, like face verification~\cite{thorpe1996speed,taigman2014deepface}, training recognizers for rare case~\cite{li2017zero,zhou2018sparc}, speaker recognition~\cite{li2020automatic}, relationship extraction and classification~\cite{rocktaschel2015injecting,levy2017zero,ye2019multi}, and knowledge graph completion~\cite{zhang2020relation,wang2019tackling}, etc. 

As mentioned in Sec~\ref{sec1}, LSL can be divided into two major branches based on the amount of training data that can be obtained: FSL and ZSL. We provide the definition of both through derivations of typical supervised ML:

\begin{definition}[\textbf{Supervised Machine Learning}~\cite{russell2002artificial,mohri2018foundations}]
is the machine learning task of learning a function $f(\cdot)$ that maps an input $x$ to an output $y$ based on example input-output pairs $D_{train} = \left\{(x_i, y_i) \right\}_{i=1}^{N}$. To get the objective pattern: $y \leftarrow f(x)$, the empirical risk $\mathcal{L}\left(f(x_{i}), y_{i} \right)$ is used to approximate expected risk $\mathcal{L}\left(f(x), y \right)$.
\end{definition}

The large scale $N$ of $D_{train}$ makes $\mathcal{L}\left(f(x), y \right) \approx \mathcal{L}\left(f(x_i), y_i \right) $ reasonable. FSL was built for situations in which $N$ is not large enough.

\begin{definition}[\textbf{Few-shot Learning}]
shares the same objective pattern $y \leftarrow f(x)$ as supervised learning, however with few training examples $D_{train} = \left\{(x_i, y_i) \right\}_{i=1}^{m}$, where $m \ll N$. Given the small $m$, it is impossible to approximate expected risk $\mathcal{L}\left(f(x), y \right)$ from the empirical risk $\mathcal{L}\left(f(x_i), y_i \right)$.
\end{definition}

Although some research attempts to solve the challenge by generating more training data~\cite{miller2000learning,hariharan2017low,antoniou2019assume}, the $C$-way $K$-shot mode is currently the mainstream paradigm of FSL: in the training set, there are $C$ prediction cases (can be classes, detection frames, or any target representation), and each case only has $K$ annotated examples, where $C \times K = m$. In such condition, the FSL system is required to learn a function $F(\cdot)$, the outcome of function $F(\cdot)$ is a model $f(\cdot)$ which can obtain the objective pattern $y \leftarrow f(x)$ swiftly with the $K$-shot scale of training data. This is another narrow definition of FSL under $C$-way $K$-shot setup, Fig.~\ref{fig2-1} illustrates the comparison of typical ML and FSL. Simultaneously, we introduce a special case of FSL:

\begin{definition}[\textbf{One-shot Learning}]
is a special case of \textbf{Few-shot Learning} that each prediction case only owns one training example, where the training examples $D_{train} = \left\{(x_i, y_i) \right\}_{i=1}^{m}, m = C \times 1$.
\end{definition}

For circumstances in which we know that the data, $m$ does not contain a sample of all possible classes (or detection frames, target representations, etc.), researchers use additional structured information for zero-sample-based learning:

\begin{definition}[\textbf{Zero-shot learning}]
contains two datasets. The first is a seen dataset $D_{\mathcal{S}} = \left\{ (x_i^{\mathcal{S}}, y_i^{\mathcal{S}}) \right\}_{i=1}^{N^{\mathcal{S}}}$, $D_{\mathcal{S}}$ contains $N^{\mathcal{S}}$ samples for training and $y_i^{\mathcal{S}} \in \mathcal{Y}^{\mathcal{S}}$. Another is a unseen dataset $D_{\mathcal{U}} = \left\{ (x_i^{\mathcal{U}}, y_i^{\mathcal{U}}) \right\}_{i=1}^{N^{\mathcal{U}}}$, which contains $N^{\mathcal{U}}$ samples only for testing and $y_i^{\mathcal{U}} \in \mathcal{Y}^{\mathcal{U}}$. $\mathcal{Y}^{\mathcal{S}}$ and $\mathcal{Y}^{\mathcal{U}}$ are the sets of seen and unseen category labels, the most important is that $\mathcal{Y}^{\mathcal{S}} \cap \mathcal{Y}^{\mathcal{U}} = \emptyset$.
\end{definition}

In ZSL, the training data $D_{\mathcal{S}}$ cannot directly provide any supervision information for the unseen test categories $\mathcal{Y}^{\mathcal{U}}$, thus $\mathcal{Y}^{\mathcal{S}}$ and $\mathcal{Y}^{\mathcal{U}}$ share a semantic attribute space $\forall y_{i} \in \left( \mathcal{Y}^{S} \cup \mathcal{Y}^{U} \right), \ \exists <a_{1}, a_{2}, \dots, a_{m}>$, which is the only bridge between seen and unseen cases and it provide supplemental reasoning description for all labels. Therefore, ZSL is also called "attribute-based learning"~\cite{lampert2013attribute}.

ZSL has two task settings with different difficulty.
\begin{enumerate}
    \item \textbf{Conventional ZSL (CZSL)}. Test samples are all from unseen categories $\mathcal{Y}^{\mathcal{U}}$, and the ZSL model knows this prior condition before performing the prediction~\cite{lampert2013attribute}.
    \item \textbf{Generalized ZSL (GZSL)}. Test samples are from both seen and unseen categories $\mathcal{Y}^{\mathcal{U}}$ and $\mathcal{Y}^{\mathcal{S}}$. Therefore, the ZSL model needs to determine whether the sample is from an unseen class, and also predict the category of the sample~\cite{xian2018zero}.
\end{enumerate}

The use of LSL, as opposed to other supervised learning approaches, is motivated by the desire to:
\begin{itemize}
\item \textit{Alleviate machine learning's dependence on data.} Large-scale data training is a difficult to aquiree. Not only do researchers suffer from the labor cost of collecting data, but it must be annotated and curated. Depending on how they have designed collection, the data can contain artificial bias~\cite{wolpert1997no}. The reliance on data makes it difficult for typical ML models to be used in related, but different tasks. Additionally, privacy concerns, such as those for medical data~\cite{patil2014big,cios2002uniqueness} make collecting large datasets difficult. Finally, some data is rare, like novel molecular structures~\cite{altae2017low}
or precious art~\cite{koniusz2018museum}, limiting the data available.
\item \textit{Expanding the generalization capabilities of machine learning.} Generalization is the ultimate goal pursued by all ML algorithms~\cite{zhang2016understanding,kawaguchi2017generalization,bousquet2002stability}. As we mentioned above, the typical ML models must be bound to the data they encounter, they cannot quickly adapt to different tasks and can be easily wrecked by well-designed attacks~\cite{finlayson2019adversarial}. With LSL, the $F(\cdot)$ in FSL gives the model $f(\cdot)$ highly generalized definition that is independent of specific tasks, allowing the model to be quickly trained on new tasks. ZSL implements unsupervised task transfer through the representation of high-level structured knowledge.
\item \textit{Apply human-like analogy and reasoning.} Humans can identify a stranger based on a few images, or they can recognize new species that have never been seen through a brief description.  Humans can also imagine completely illusory things through the materials that exist in reality, which is discussed in some pioneering studies of LSL~\cite{lake2015human}.
\end{itemize}

Structured knowledge has played a beneficial role in the process of LSL, and is necessary for some situations. For instance, KG can intuitively provide information for the few-shot KG reasoning~\cite{wang2019meta} and graph classification~\cite{chauhan2020few}. The pre-built KG can help the LSL image recognition system to model the category relationship~\cite{chen2020knowledge,wang2018zero,kampffmeyer2019rethinking}. Moreover, in ZSL, the semantic attribute space is required for the technique and is a type of structured knowledge. It is usually constructed by experts with a small manual amount and can be in various forms, such as attribute distribution space~\cite{farhadi2009describing,ferrari2008learning,parikh2011relative,lampert2009learning}, word embedding~\cite{mikolov2013efficient,pennington2014glove,frome2013devise}, or description text embedding~\cite{reed2016learning}.

\section{General Low-shot Learning Models}
\label{sec3}

 Based on these definitions and requirements presented in Section~\ref{sec2}, this section briefly introduces the popular models of LSL before formally reviewing LSL in structure data modeling.

\subsection{Few-shot Learning}
When the developer can obtain a small number of training examples, the FSL models are dominated by the following strategies:

\textbf{Methods based on augmenting the data resource:} Directly increasing the amount of training data is the most intuitive idea. Some methods transform the existing training set data and assign pseudo-labels to add new data to the training set~\cite{miller2000learning,hariharan2017low,antoniou2019assume}. Another line of strategies~\cite{douze2018low,li2019learning,zhou2018sparc} give high-confidence pseudo-labels to unlabeled data sets to turn them into training data. Also, the similar data sets can be used as the source to generate data into the few-shot training set~\cite{tsai2017improving,gao2018low}. Strictly speaking, this method does not fall within the FSL paradigm. It also has the following limitations: the simple transformation of replicated data is weak in diversity; there is no similar data set; and the initial classifier cannot give reliable pseudo labels.

\textbf{Methods based on learning of analogy tasks:} This strategy is to acquire knowledge from other data-rich tasks to help the few-shot task execution. For example, we need to classify dog breeds, but we have only a few examples, and we can get some experience from bird classification or cat classification tasks. This series of methods are usually constructed by on FSL task and several similar data-rich tasks. Some of them share all or part of the parameters, and allow the FSL model to fine-tune the parameters on sufficient data task to quickly obtain feature modeling capabilities~\cite{hu2018few,sun2019meta,benaim2018one}. Some methods bind FSL tasks and regular tasks loosely, encouraging FSL models to learn similar weights to other tasks~\cite{yan2015multi,luo2017label}. Some studies also apply the Bayes-based models or other generative models to extract the prior knowledge from other data sets, thus enhance the feature modeling~\cite{fei2006one,lake2015human,salakhutdinov2012one,edwards2016towards}.

\textbf{Metric-based methods:} FSL models based on metric learning usually require all samples to be mapped into an embedding space. They shift the emphasis of learning from predicting labels to predicting the distance between samples. This makes the FSL task  similar to an information retrieval problem during testing~\cite{triantafillou2017few}. In the process of metric modeling, the researchers chose various encoding methods, like the Siamese network that receives the pairwise samples input with model parameters shared~\cite{koch2015siamese,droghini2018few}, other series of methods embed training samples collectively with bidirectional LSTM~\cite{vinyals2016matching}, attentional LSTM~\cite{salakhutdinov2011learning}, CNN~\cite{sung2018learning}, or Graph Neural Network ~\cite{kim2019edge}, some incorporate task-specific information~\cite{oreshkin2018tadam,zhao2018dynamic}. When inquiring, Prototypical Networks (ProtoNet) proposed to calculate the distance between the test sample and each cluster center of training samples~\cite{snell2017prototypical}. Another line of studies attempt to extract embedded information into external memory with the form of key-value, thereby compressing the cost of representation and parameters~\cite{sukhbaatar2015end,santoro2016meta,munkhdalai2017meta,cai2018memory}.

\begin{SCfigure}[][h]
\centering
\includegraphics[scale=0.5]{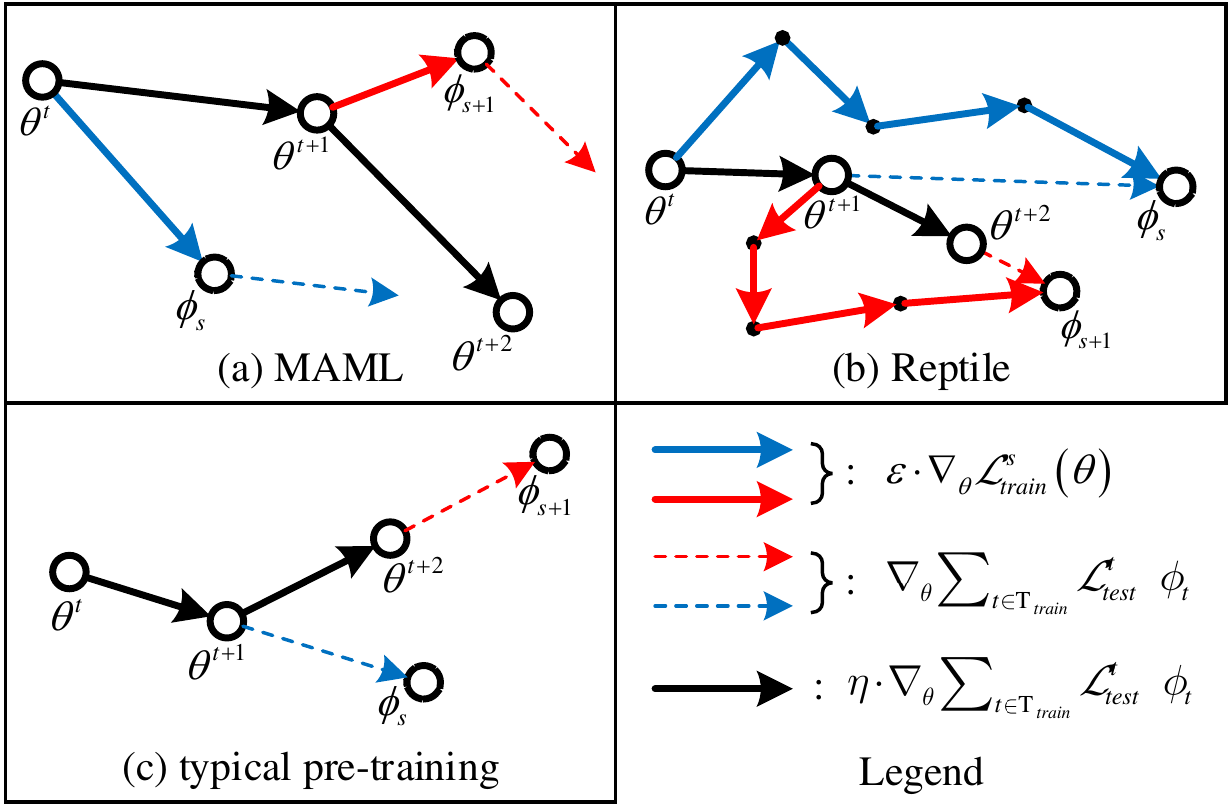}
\caption{Comparison of the meta-learning methods and the ordinary pre-training. (a) MAML, which updates once on a sampled training task and optimizes the meta-learner $\theta$ with the predicted performance of all training tasks; (b) Reptile, which updates $\theta$ follow multi-step optimization on the training task; The optimization of meta-learner focuses more on the future learning potential of the model. (c) The typical pre-training, which updates parameters according to the current optimal direction.}
\label{fig3-1}
\end{SCfigure}

\textbf{Methods based on meta-learned parameters:} Some literature defines FSL as meta-learning in a narrow sense, based on the commonality between learning tasks in FSL.  As a classic paradigm of FSL introduced in the Section~\ref{sec2}, meta-learned parameter methods use the tasks as the learning objects, the output of which is an FSL model with good  initialization~\cite{lee2018gradient,finn2018probabilistic}. There are several examples. Model-Agnostic Meta-Learning (MAML)~\cite{finn2017model}, sets the goal as learning a meta-learner $\theta$. To do this,  it will sample a set $\mathrm{T}_{train}$ of training tasks, and investigate the loss $\mathcal{L}_{test}(\phi_t)$ of the model on the test set of each tasks $t$, and then update $\theta$ as: $\theta \leftarrow \theta - \eta {\nabla}_{\theta} \sum_{t \in \mathrm{T}_{train}} \mathcal{L}_{test}^{t}(\phi_t)$. When facing a specific task $s$, meta-learner $\theta$ is used to optimize the FSL model parameter $\phi_{s}$ as: $\phi_{s} = \theta - \varepsilon \nabla_{\theta} \mathcal{L}_{train}^{s}(\theta)$. Reptile~\cite{nichol2018first} observes the destination of multi-step updates of $\theta$ on each training task, and serves as the update of $\phi_{s}$. Fig.~\ref{fig3-1} shows the comparison of MAML, Reptile, and typical pre-training.

\textbf{Methods based on memory enhanced optimizer:} Based on the meta-learned methods above,  \cite{andrychowicz2016learning,ravi2016optimization} cleverly proposed to use the status of LSTM to store the update process of $\phi_{s}$, which acts the role of meta-learner $\theta$, plus learning LSTM itself, and finally realize learning to learn. This is also beneficial to record the updated memory of $\phi_{s}$.

\subsection{Zero-shot Learning}
When we face new prediction targets without any training data, ZSL gives a more unified strategy than FSL. According to the definition given in Section~\ref{sec2}, the ZSL model $F_{\theta} \left(\cdot, W_{\theta} \right)$ first needs to predict the attributes $\mathcal{A}_{i} = \left[a_1, a_2, \dots, a_{m_i} \right]$ belong to the input sample $x_{i}$ as: $\mathrm{a} = F_{\theta} \left(x_{i}, W_{\theta} \right)$, then obtain the prediction result $y_{i}$ based on attribute reasoning: $y_{i} = F_{\phi} \left(\mathcal{A}_{i} \right)$. Since the seen category and the unseen category share an attribute space, when training, the ZSL models are usually dedicated to learning the mapping of samples to attributes. In order to better complete attribute modeling and reasoning, the ZSL models have used the following methods:

\textbf{Semantic Inference Methods} Since ZSL was first proposed, the earlier ZSL models~\cite{lampert2009learning,lampert2013attribute} have divided the execution of ZSL into two steps: 1. Attribute prediction; 2. Semantic inference. The first step is to train the model to output the attributes corresponding to the samples, and the second step is to predict the category based on these attributes. Even with external multi-modal knowledge assistance\cite{socher2013zero}, the semantic inference methods still have apparent problems: 1. Based on inaccurate assumptions that attribute labels are independent of each other~\cite{romera2015embarrassingly}; 2. a gap between the prediction of attribute labels and the category objective, resulting in the phenomenon of "domain shift"~\cite{fu2015transductive}.

\textbf{Semantic Mapping Methods} In response to the above two problems of the semantic inference methods, researchers created a  process~\cite{romera2015embarrassingly,zhang2017learning} to directly map semantic attributes, using various techniques such as bidirectional mapping~\cite{akata2013label,liu2018generalized}, depth encoding~\cite{morgado2017semantically}, auto-encoder~\cite{kodirov2017semantic}, and joint mapping~\cite{akata2015evaluation,ye2019progressive}. These methods not only ease the problem of ¡°domain shift¡±, but also greatly simplifies the ZSL process. Almost all followed ZSL models adopt the semantic mapping strategy.

\textbf{Feature Generation Methods:} These methods use the generative models like generative adversarial network (GAN)~\cite{huang2019generative,elhoseiny2019creativity,yu2020episode} or  variational auto-encoder (VAE)~\cite{xian2019f,gao2020zero} to generate features with an attribute as the pivot. It improves the coupling of the channel of ¡°feature-attribute-prediction¡±.

\textbf{Attribute Space Extension Methods:} Some researchers believe that pre-defined attributes are incomplete and cannot adequately describe the categories and samples. Therefore, designing external feature spaces helps the model to represent implicit attributes~\cite{li2018discriminative,liu2019attribute,zhu2019generalized}.

\textbf{Transductive Methods:} This type of method is similar to transfer learning. They allow the model to access unlabeled unseen data~\cite{fu2015transductive,song2018transductive,xu2017transductive}, thereby helping the model to establish the transduction association from the seen category to unseen.

\textbf{Knowledge Graph-Assisted Methods:} In addition to the above methods, structured knowledge is also used to help model the semantic relationship between categories and attributes. These works usually use self-built topologies~\cite{gong2019zero,chen2020zero} or external ontology libraries~\cite{wang2018zero,kampffmeyer2019rethinking,liu2018combining} as structured auxiliary information to promote the relationship representation of the ZSL model.

In addition to the above methods, the ZSL models also introduce various tricks, such as attention mechanism~\cite{liu2019attribute}, manifold space modeling~\cite{fu2017zero} or  regularized bias loss~\cite{song2018transductive}. The modeling of structured data can form a benign closed loop with LSL.  In Sections~\ref{sec4} and \ref{sec5}, we will introduce the related researches of LSL on the modeling of structured data and the benefits from structured data mining feed to LSL, respectively.

\section{Low-shot Learning on Structured Data}
\label{sec4}

In this section, we look specifically at Low-shot learning on structured data. We break the discussion into two parts: Structured Data Modeling with LSL and Using Structured Data in LSL. We will introduce the application of LSL with respect to different structured data modeling tasks, including: entity/node identifying, relation classification and extraction, graph/ontology modeling, query/question understanding, and some other structural modeling tasks, etc. We will explain some special handling of LSL methods for structured data, main challenges, and the adaptive changes made to the original LSL model. We then move on to describe the use of structured data in LSL techniques.

\subsection{Structured Data Modeling with LSL}
\label{sec4.1}

Although LSL was proposed and initially verified in the field of computer vision, the benefits of its advanced nature are not limited to the area of visual modeling. As a general-purpose machine learning strategy, various algorithms of LSL can be generalized and applied to machine learning tasks for the various data types, including  structured data modeling. Since LSL started  later on structured data tasks than visual tasks, most LSL structured data modeling methods follow the  strategies that were successful in the field of vision.

\subsubsection{Entity/Node Identifying}
\label{sec4.1.1}

Nodes are the most basic constituent elements in a structured knowledge graph.  Nodes usually refer entities in the real world. Entity/node modeling tasks such as recognition and linking, with their unique task characteristics, map naturally to the reasons low-shot solutions are applied. Entities map the diversity of real-world objects, and large-scale labeling of all entities is unrealistic, not to mention the problems of rare domains and cross-language labelling. Additionally, new concepts will continue to emerge over time. Thus any model developed must be able to identify novel, as yet unseen entities. Finally, structured labelling requried for some  multi-step entity identification tasks, such as named entity recognition (NER) and entity linking (EL) create large labor costs and extremely poor portability.

Based on these requirements, we describe how LSL-based entity modeling has been used in diverse entity and node identification scenarios.

\begin{wrapfigure}{l}{0.5\textwidth} \vspace{-5pt}
\centering
\includegraphics[width=0.5\textwidth]{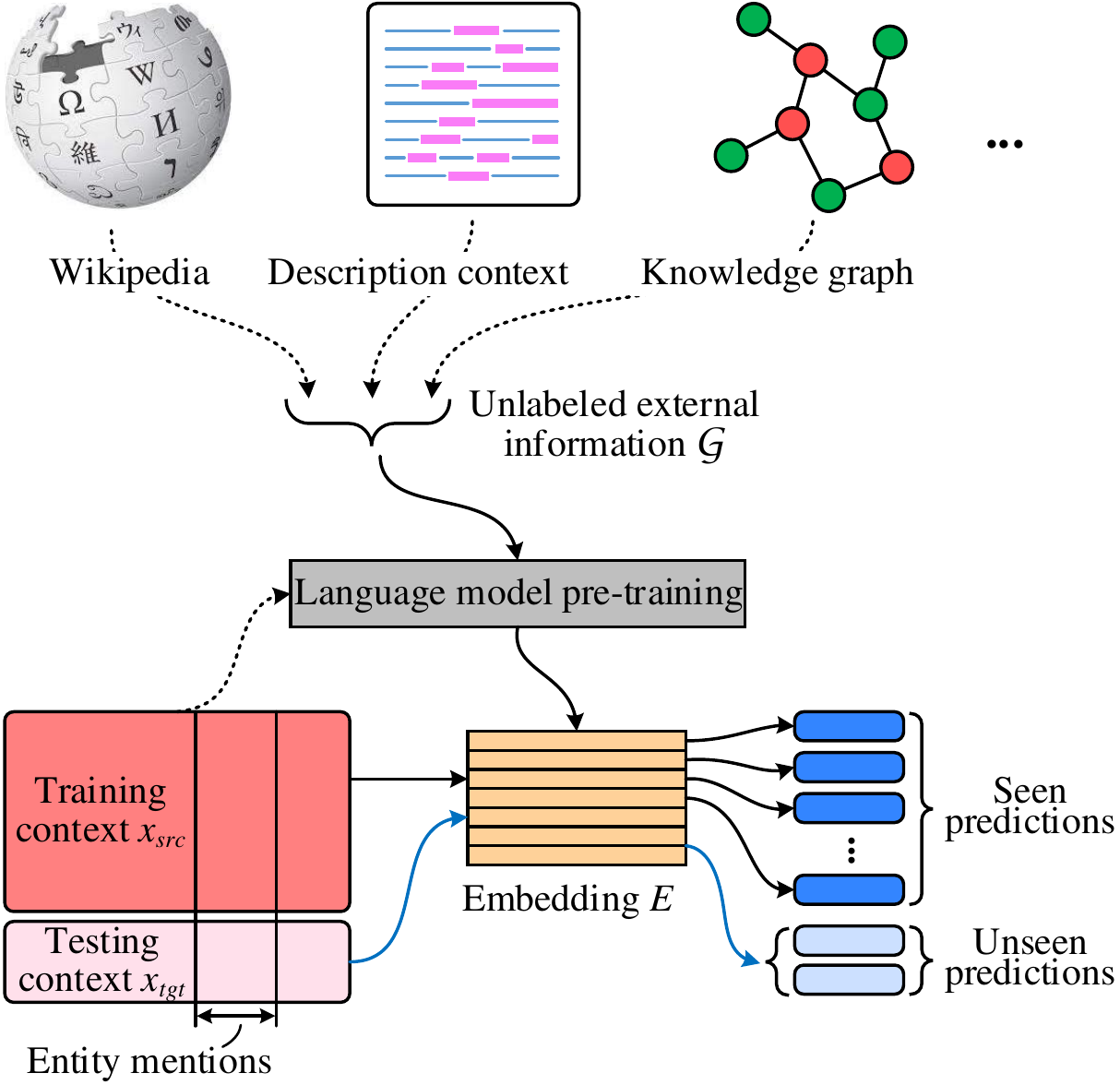}
\caption{Illustration of ZSL entity identifying with pre-trained knowledge embedding. Training embedding $E$ based on easily accessible external knowledge and corpus to help the generalization of entity concepts, where the solid line represents the required data flow, the dotted line is optional.} \vspace{-5pt}
\label{fig4-1-1-1}
\end{wrapfigure}

\uppercase\expandafter{\romannumeral1}. \textbf{Coherent zero-shot training}. Although there is an urgent need for zero-shot generalization formulation in entity modeling due to the dilemmas mentioned above, researchers take advantage of natural language processing (NLP) to make the domain generalization and transformation of entity identifying more convenient. The generalized semantic system of human language itself provides a naturally shared space $E$ for different entity types. For example, an embedding $E (\mathcal{G})$ pre-trained on unlabeled information $\mathcal{G}$, like Wikipedia, description context, and any available knowledge graph, has the generalization function of concept transformation from seen entities $x_{src}$ to unseen entities $x_{tgt}$. As illustrated by Fig.~\ref{fig4-1-1-1}, unlike general ZSL models, which are limited to manually defined attribute spaces, embedding $E$ can access all open-scope offline resources, thereby achieving "utilizing knowledge to learn knowledge". In Reference~\cite{sil2012linking,qu2016named}, transfer learning is applied to express the generalized correlation from the source domain to a novel domain. Similarly, in~\cite{sil2012linking}, an approach is proposed to conduct domain adaptation from a distant supervision model. This approach models the transformation to any new domain generically by re-formalize the domain representation. \cite{wang2015language} proposes to fully leverages the knowledge base structure, in which the linked structures in KBs such as DBpedia~\footnote{http://wiki.dbpedia.org} are mainly considered to evaluate entity candidates, and it jointly validates entities with source context and multi-ranking results. \cite{huang2018zero} learns a shared metric space for event mentions and event types with the target ontology.

To further reduce the additional operations performed when transferring to the new domains, pre-trained embeddings can be used to extend to unseen types~\cite{huang2016building,ma2016label,yuan2018otyper,obeidat2019description,imrattanatrai2019identifying}.  Among them, \cite{huang2016building} learns the embedding with both linguistic context and knowledge base, then utilizes hierarchical clustering to gather new entities. Meanwhile,  \cite{ma2016label} adapt zero-shot learning on a context prototype and WordNet~\cite{miller1995wordnet} hierarchy embeddings. \cite{yuan2018otyper} builds the word embedding from text resources and projects the embeddings into a common space. Wikipedia descriptions of entity types are used by  \cite{obeidat2019description} to learn the medium space. In \cite{imrattanatrai2019identifying},  selected the components of the knowledge graph are used as the embedded object, with the structured embedding of the knowledge graph used as the entity property space. \cite{wu2019zero} utilizes a bi-encoder to build a dense space based on the semantic model and carries out entity retrieval on this space. Recently, attention encoding on entity types is employed to map context semantic to types~\cite{ren2020neural}.

\uppercase\expandafter{\romannumeral2}. \textbf{Progressive zero-shot training}. Most of the training strategies described above maintain a coherent process, and we can roughly formulate them as: $y = F_{[\sim]} \left(x, E(\mathcal{G}) \right)$. Another line of studies uses multi-stage training to achieve more precise aid from structured information. Some techniques require a pivot result, which can more easily receive assistance from external knowledge and lead to the final prediction, as:
\begin{equation}
\label{eq4-1-1-1}
s = F_{[\sim, s]} (x) \longrightarrow y = F_{[s, \sim]} \left(s, E(\mathcal{G}) \right).
\end{equation}

The system described in \cite{pasupat2014zero}  generates a set of extraction predicates, then extracts entities from this set with structured background. \cite{zhou2018zero} ground a given mention to a set of type-compatible Wikipedia entries and then infer the target mention's types with multiple offline resources.

On the other hand, some studies prefer to introduce the support of structured knowledge in each stage of entity identification, as:
\begin{equation}
\label{eq4-1-1-2}
s = F_{[\sim, s]} \left(x, E(\mathcal{G}) \right) \longrightarrow y = F_{[s, \sim]} \left(s, E(\mathcal{G}) \right).
\end{equation}

For instance, \cite{guerini2018toward} first trains a gazetteer on entity names then apply a neural classifier to a user utterance to rank the entity candidates. Both stages utilize the embedding model trained on Wikipedia corpora. \cite{logeswaran2019zero} proposes domain adaptive pre-training (DAP), where the model is pre-trained only on the target-domain data, and the strong reading comprehension function is used to generalize to unseen entities. In  \cite{liu2020zero}, a model is constructed as a multi-task framework: $\mathrm{s} = \{s_{1}, s_{2}, \dots, s_{n} \}$.  $E(\mathcal{G})$ is used to regulate the weights of sub-task elements in $\mathrm{s}$ as $F_{[\mathrm{s}, \sim]}  = \mathrm{s} \circ F\left(E(\mathcal{G}) \right)$, where $\circ$ refers to the element-wise multiplication.

Finally, knowledge can be used to build the embedding space $E$, and simultaneously find a pivot in $E$ as the transition for searching final results. This can be formulated as:
\begin{equation}
\label{eq4-1-1-3}
E, s = F_{[\sim, s]} (x, \mathcal{G}) \longrightarrow y = F_{[s, \sim]} \left(E(s, \mathcal{G}) \right).
\end{equation}

This technique has been applied  to solve the cross-language or cross-domain challenges in the entity linking (EL) task, use a highly correlated high-resource language to establish the embedding space, and find a pivot object in it to perform the final prediction to the target entity \cite{rijhwani2019zero}. This method also involves the metric-based strategy, which we will discuss in detail later.

\begin{table}[t]
\centering
\footnotesize
\caption{Resources required for different types of entity linking (EL) methods~\cite{logeswaran2019zero}.}
\label{tab4-1-1-1}
\begin{tabular}{c|p{35pt}|p{35pt}|p{35pt}|p{35pt}|p{35pt}|p{35pt}}
\hline
Method & In-Domain & Seen Entity Set & Small Candidate Set & Statistics & Structured Meta Data & Entity Dictionary \\
\hline
Standard EL & \checkmark & \checkmark & -- & \checkmark & \checkmark & \checkmark \\
Cross-Domain EL & -- & \checkmark & -- & \checkmark & \checkmark & \checkmark \\
Linking to Any DB~\cite{sil2012linking} & -- & -- & \checkmark & -- & \checkmark & \checkmark \\
\hline
Zero-Shot EL & -- & -- & -- & -- & -- & \checkmark \\
\hline
\end{tabular}
\end{table}

Although the above method draws on various external knowledge resources and pre-trained embeddings, these resources need no manual annotation and are easy to obtain. Additionally,  the pre-trained embeddings are offline. Depending on the entity linking (EL) task, different training materials are required for ZSL or other methods, as decribed in  Table~\ref{tab4-1-1-1}  ~\cite{logeswaran2019zero}.

\begin{table}[t]
\centering
\footnotesize
\caption{ Characteristics of LSL Methods for Entity/node Modeling \& Identifying.}
\label{tab4-1-1-2}
\begin{tabular}{p{50pt}|p{90pt}|p{130pt}|p{80pt}}
\hline
Training type & Specific task & Technology highlight & Reference \\
\hline
\multirow{8}{50pt}[-15pt]{Coherent ZSL} & Entity linking & Distant supervision \& Transfer learning & \citet{sil2012linking} \\
\cline{2-4}
 & Named entity recognition & Transfer learning & \citet{qu2016named} \\
\cline{2-4}
 & Entity linking & Representation learning on KBs & \citet{wang2015language} \\
\cline{2-4}
 & Event extraction & Word (context) embedding \& Metric learning & \citet{huang2018zero} \\
\cline{2-4}
 & Entity typing & Word (context) embedding, Clustering \& Attenytion mechanism & \citet{huang2016building,ma2016label,yuan2018otyper,ren2020neural} \\
\cline{2-4}
 & Entity typing & Description pre-training & \citet{obeidat2019description} \\
\cline{2-4}
 & Entity property identifying & KG component embedding & \citet{imrattanatrai2019identifying} \\
\cline{2-4}
 & Entity retrieval & BERT pre-training & \citet{wu2019zero} \\
\hline
\multirow{6}{50pt}[-2pt]{Progressive ZSL} & Event extraction & Structured feature generation & \citet{pasupat2014zero} \\
\cline{2-4}
 & Entity typing & Wikipedia representation learning & \citet{zhou2018zero} \\
\cline{2-4}
 & Named entity recognition & Neural gazetteer & \citet{guerini2018toward} \\
\cline{2-4}
 & Named entity recognition & Multi-task learning \& Gate mechanism & \citet{liu2020zero} \\
\cline{2-4}
 & Entity linking & Domain adaptive pre-training & \citet{logeswaran2019zero} \\
\cline{2-4}
 & Entity linking & Metric learning \& Pivot-based learning & \citet{rijhwani2019zero} \\
\hline
\multirow{9}{50pt}[-6pt]{FSL} & Category characterization & Label propagation & \citet{zhou2018sparc} \\
\cline{2-4}
 & Named entity recognition & Pre-training \& Hyper-parameter tuning & \citet{hofer2018few} \\
\cline{2-4}
 & Named entity recognition & Prototypical network & \citet{fritzler2019few} \\
\cline{2-4}
 & Entity typing & Multi-task learning & \citet{ling2019learning} \\
\cline{2-4}
 & Word embedding learning & MAML meta-learner & \citet{hu2018few} \\
\cline{2-4}
 & Node classification & MAML meta-learner & \citet{zhou2019meta} \\
\cline{2-4}
 & Node classification & Graph prototypical networks & \citet{ding2020graph} \\
\cline{2-4}
 & Event detection & Semantic similarity learning & \citet{peng2016event} \\
\cline{2-4}
 & Event detection & Prototypical network \& Memory enhanced optimizer & \citet{deng2020meta} \\
\hline
\end{tabular}
\end{table}

\uppercase\expandafter{\romannumeral3}. \textbf{Few-shot training}. When some training samples are available, many general LSL techniques are widely utilized in entity/node identifying with few training samples, thereby further enhancing the generalized learning ability of the model and reducing coupling and dependence of domain knowledge and corpus. The most intuitive design is based on pseudo label propagation~\cite{zhou2018sparc} or multi-task accompanying training~\cite{hofer2018few,ling2019learning}.  \cite{hofer2018few} utilize 4 kinds of pre-training and auxiliary knowledge embedding strategies, and apply multi-task training and  grid search to optimize hyper-parameters.   Fill-in-the-blank tasks to learn context independent representations of entities has been used \cite{ling2019learning}, as has self-learning to gradually learn label propagation from the "easy" nodes to the "difficult" nodes, with the graph contextual prediction to guide the generation of the pseudo label \cite{zhou2018sparc}. In addition, metric-based methods~\cite{peng2016event,fritzler2019few},  meta-leaner~\cite{zhou2019meta,hu2018few}, and memory enhanced methods~\cite{deng2020meta} as described previously have been applied to the entity/node identification task. For example, the  MAML~\cite{finn2017model} model has been applied to node classification in graph data \cite{zhou2019meta}. MAML was also combined with attention regression of  multiple observers \cite{hu2018few}. ProtoNet~\cite{snell2017prototypical} has been used to construct the metric space of named entities~\cite{fritzler2019few} and been upgraded to graph learning mode for few-shot node classification~\cite{ding2020graph}. In event detection, the prototypical space and memory-based multi-hop structure are used to distill contextual representation with event mentions~\cite{deng2020meta}. Meanwhile, similarity between the examples and the candidates from the event mention and event type ontology has been used in few-shot event detection \cite{peng2016event}.

 Table~\ref{tab4-1-1-2} summarizes the LSL work in entity/node identifying. It can be concluded that auxiliary information such as word embedding, NLP language model, external structured knowledge provide the implicit semantic transfer bridge for entity/node ZSL modeling. Their role analogies to the shared attribute space in image modeling ZSL. However, the understanding of context semantic provides a smoother transformation effect. The progressive model modularly splits the entity/node modeling task to use semantic knowledge more precisely~\cite{pasupat2014zero,zhou2018zero}, while the split learning may bring some risk of error accumulation (the error of stage $F_{[\sim, s]}$ is delivered to stage $F_{[s, \sim]}$). The pre-training of language embedding models is widely combined with general FSL methods based on metrics and meta-learners in FSL~\cite{huang2018zero,fritzler2019few,hu2019few,deng2020meta}. However, for extremely rare domains or languages, there still be serious domain bias in the pre-trained semantic space. There is still a lot of room for exploration in the application of metrics-based, meta-learning, such technologies for entity/node modeling.

\subsubsection{Relation Classification \& Extraction}
\label{sec4.1.2}

Relation extraction \& classification (RE/C) is a crucial independent task in knowledge graph and ontology modeling. The RE/C task can be regarded as a classification task for text: what kind of relationship exists between two specific entities $(x_{i}, x_{j})$ which are contained in the text (usually sentences, sometimes long texts). Compared with the CV tasks, RE/C has more urgent practical demands for LSL technology: 1. Relations are how humans define the connection between things, reflecting the diversity of human thinking and always updating. The ideal RE/C model should not be limited by the scope of the known relation library and must face various novel relations; 2. The relation types follow the long-tail distribution, rare relations have sufficient training examples, while relations with few examples account for the vast majority.

\begin{wrapfigure}{r}{0.4\textwidth} \vspace{-5pt}
\centering
\includegraphics[width=0.4\textwidth]{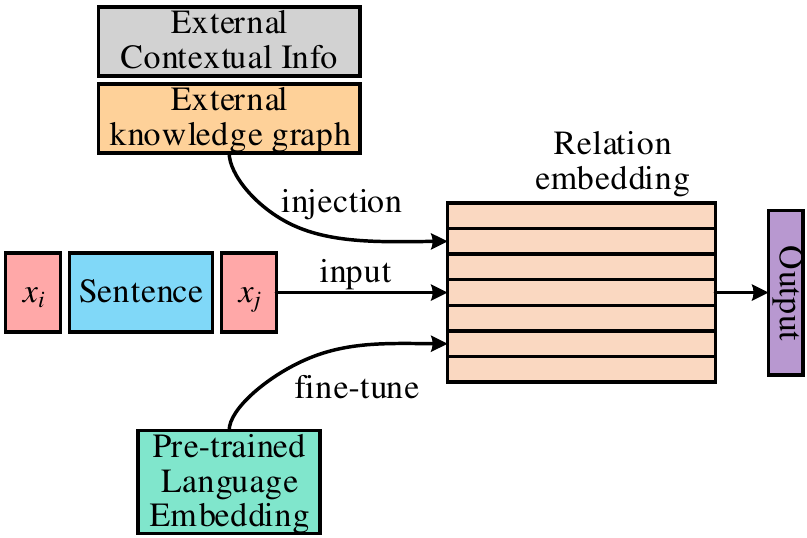}
\caption{The prior knowledge assisted LSL relation extraction/classification.} \vspace{-5pt}
\label{fig4-1-2-1}
\end{wrapfigure}

\uppercase\expandafter{\romannumeral1}. Confronted with very few training samples,  researchers intuitively aim to fill the holes in the training data \textbf{with the prior knowledge}. Although distant supervision technology~\cite{mintz2009distant} uses external texts to supplement the small number of training samples in the knowledge graph, RE/C training materials are still limited to the relation types in the knowledge base, and the external texts also contain a lot of noise. The external knowledge-assisted method can help the models to complement themselves from other knowledge sources by injecting more information into the matrix factorization or relation embedding process (Fig.~\ref{fig4-1-2-1}). Some studies~\cite{rocktaschel2015injecting,demeester2016lifted} utilize knowledge graphs in other related fields or generalized ontologies, like WordNet~\cite{miller1995wordnet} or PPDB~\cite{ganitkevitch2013ppdb}, to help models supplement relational materials. Context patterns can be learned from external text has formed the novel relation pairs~\cite{obamuyide2017contextual}. Additionally,  the \textbf{pre-trained} word embeddings from a large-scale language model (such as BERT~\cite{devlin2019bert}), can filter the precise relation pairs \cite{papanikolaou2019deep}.

\uppercase\expandafter{\romannumeral2}. In order to further solve the problem of lack of training data, some studies combine the prior knowledge-assisted technology with \textbf{semi-supervised learning} to help the model establish a transductive path from a data-rich source to a data-lacking target.  Graph convolutional networks (GCN) to model the relational knowledge and provide coarse-to-fine modeling for the external information with knowledge-aware attention have been used \cite{zhang2019long}. Alternatively, heterogeneous graphs based on the message passing between entity nodes and sentence nodes create a set of relationships, to which adversarial learning is applied to alleviate data noise \cite{xie2020heterogeneous}.

\begin{SCfigure}[][h]
\centering
\includegraphics[scale=0.8]{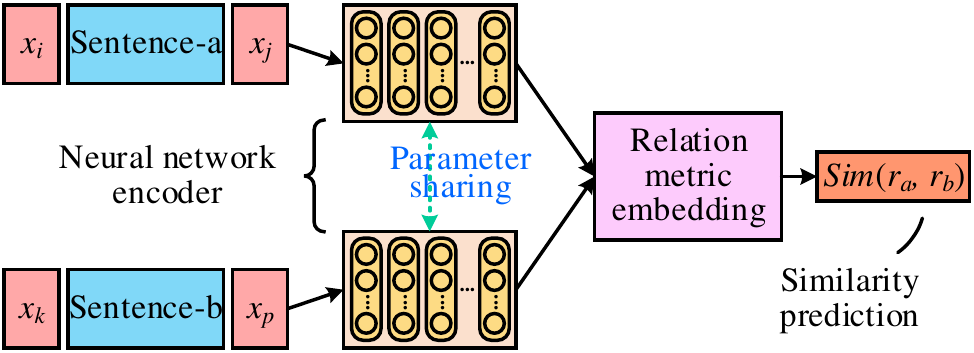}
\caption{Metric-based LSL relation extraction/classification. The input is the comparative relation pair, through the modeling of the relational metric, output the similarity between relations. The neural network encoder that models each relation can be the siamese network with parameter sharing.}
\label{fig4-1-2-2}
\end{SCfigure}

\uppercase\expandafter{\romannumeral3}. \textbf{Metric-based} LSL model is widely utilized in RE/C learning with few training samples (Fig.~\ref{fig4-1-2-2}). Most of them apply the classic metric-based LSL method used in the CV field. \cite{yuan2017one} and \cite{wu2019open} propose to use siamese convolutional network to build a relation metric model. Virtual adversarial training over unsupervised models have been used to improve the distance prediction \cite{wu2019open}. Meanwhile, \cite{xiong2018one} utilizes the matching network~\cite{vinyals2016matching} to discover similar triples to a given reference triple. In similar work, an unsupervised model creates an agnostic relation representation, and then uses it for metric modeling between relations, where the pre-trained language model is introduced to encode relation sentences \cite{soares2019matching}. The ProtoNet~\cite{snell2017prototypical} is  popular in metric-based LSL RE/C tasks~\cite{gao2019hybrid,fan2019large,ye2019multi} including designs multi-level attention schemes to highlight the crucial instances and features respectively \cite{gao2019hybrid}, adopting the large-margin ProtoNet with fine-grained sentence features \cite{fan2019large}, integrating external descriptions to enhance the origin ProtoNet~\cite{yang2020enhance}, and using a multi-level structure to encode the metric prototype through various stage \cite{ye2019multi}. Metric-based LSL has also been  integrated with semi-supervised learning, exploring novel relations from the seed relation set with a relational metric and snowball strategy \cite{gao2020neural}.

\begin{wrapfigure}{r}{0.4\textwidth} \vspace{-5pt}
\centering
\includegraphics[width=0.4\textwidth]{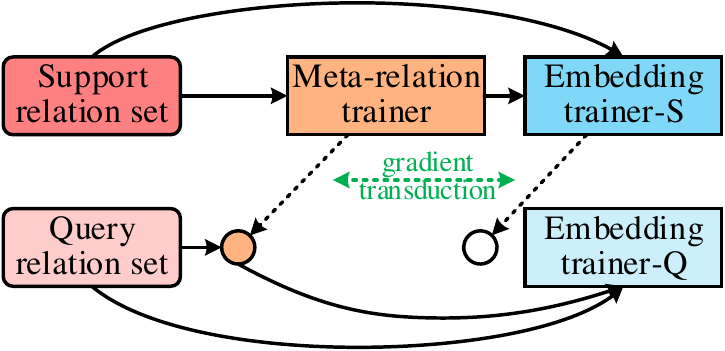}
\caption{A classic example of the gradient-based meta-learner for LSL relation extraction/classification. The training task set is called the support set, and the test task set is the query set. The relation-meta trainer is used to learn the higher-order representation of the relations, and updates the initialization on the query set with the embedding gradient from the support set.} \vspace{-5pt}
\label{fig4-1-2-3}
\end{wrapfigure}

\uppercase\expandafter{\romannumeral4}. \textbf{Gradient-based meta-learners} attempt to generalize the learning experience of the relation classifier to achieve speedy RE/C training with few training examples. As shown in Fig.~\ref{fig4-1-2-3}, a classic example of RE/C meta-learner~\cite{chen2019meta} utilizes the gradient from embedding module to optimize the relation-specific meta information to make model learn the most important knowledge and learn faster. \cite{obamuyide2019meta} use Reptile~\cite{nichol2018first} meta learning algorithm to realize few-shot relation extraction. \cite{obamuyide2019model} applies another typical MAML~\cite{finn2017model} meta-learner to construct the limited supervised, \cite{bose2019meta} further utilizes the MAML algorithm to propose a meta graph neural network (GNN) model and carries out efficient initialization of the link prediction trainer on the graph structure. \cite{baek2020learning} proposes a novel transductive meta-learning framework called Graph Extrapolation Networks. which meta-learns both the node embedding network for inductive inference (seen-to-unseen) and the link prediction network for transductive inference (unseen-to-unseen). This framework generalizes the relational learning of seen-unseen entities, a larger expansion of the unseen-pseudo unseen entities is completed. \cite{qu2020few} learns the prototype vectors and the global relation graph to extract the few-shot relations in the novel domain and improve the usage of low training resources, it also proposes stochastic gradient Langevin dynamics based on MAML~\cite{finn2017model} to handle the uncertainty of the prototype vectors.

\uppercase\expandafter{\romannumeral5}. \textbf{Zero-shot} formulation is also introduced to RE/C methods to make full use of semantic information to realize the identification of brand new relations without any training data. Similar to the conventional ZSL paradigm, ZSL's RE/C still requires a task to help the model conduct semantic migration. \cite{levy2017zero} associates one or more natural-language questions $q_{x}$ with answers $y$ to each relation type $R(x, y)$, which translates the ZSL RE/C problem to a pre-designed reading comprehension task. \cite{obamuyide2018zero} preforms zero-shot relation classification by exploiting relation descriptions. The RE/C task is formulated as a textual entailment problem~\cite{dagan2005pascal}. \cite{qin2020generative} also learns the semantic features of relations from their text description, it proposes to leverage GAN to establish the connection between text and knowledge graph domain.

\begin{table}[t]
\centering
\footnotesize
\caption{Characteristics of LSL Methods for Relation Extraction \& Classification.}
\label{tab4-1-2-1}
\begin{tabular}{p{50pt}|p{50pt}|p{70pt}|p{80pt}|p{80pt}}
\hline
Method & Type & Specific task & Technology & Reference \\
\hline
\multirow{4}{50pt}[-10pt]{Based on auxiliary prior knowledge} & FSL & Relation extraction & \multirow{2}{80pt}{External knowledge graph injection} & \citet{rocktaschel2015injecting} \\
\cline{2-3} \cline{5-5}
 & FSL \& ZSL & Relation embedding &  & \citet{demeester2016lifted} \\
\cline{2-5}
 & Unsupervised & Relation extraction & Pre-trained language model fine-tuning & \citet{papanikolaou2019deep} \\
\cline{2-5}
 & FSL (OSL) & Relation extraction & External contextual pattern injection & \citet{obamuyide2017contextual} \\
\hline
\multirow{2}{50pt}{Semi-supervised} & FSL & Relation extraction & \multirow{2}{80pt}{GNN} & \citet{zhang2019long} \\
\cline{2-3} \cline{5-5}
 & FSL & Relation classification &  & \citet{xie2020heterogeneous} \\
\hline
\multirow{4}{50pt}[-20pt]{Metric-based framework} & FSL (OSL) & Relation extraction &  Siamese convolutional network & \citet{yuan2017one,wu2019open} \\
\cline{2-5}
 & FSL (OSL) & Relation extraction & Matching network & \citet{xiong2018one,soares2019matching} \\
\cline{2-5}
 & FSL & Relation classification & Prototypical networks & \citet{gao2019hybrid,fan2019large,ye2019multi,yang2020enhance} \\
\cline{2-5}
 & FSL & Relation extraction & Relational siamese network \& Neural snowball & \citet{gao2020neural} \\
\hline
\multirow{6}{50pt}[-25pt]{Gradient-based meta-learners} & FSL & Link prediction & Relation-specific gradient optimization & \citet{chen2019meta} \\
\cline{2-5}
 & FSL \& Lifelong learning & Relation extraction & Reptile meta-learner & \citet{obamuyide2019meta} \\
\cline{2-5}
 & FSL & Relation extraction & Bayesian meta-learner & \citet{qu2020few} \\
\cline{2-5}
 & FSL & Relation classification & MAML meta-learner & \citet{obamuyide2019model} \\
\cline{2-5}
 & FSL & Link prediction & MAML meta-learner \& GNN & \citet{bose2019meta} \\
\cline{2-5}
 & FSL & Link prediction & Transductive meta-learner & \citet{baek2020learning} \\
\hline
\multirow{3}{50pt}[-5pt]{Medium-task-based semantic transform} & ZSL & Relation extraction & Reading comprehension & \citet{levy2017zero} \\
\cline{2-5}
 & ZSL & Relation classification & Textual entailment & \citet{obamuyide2018zero} \\
\cline{2-5}
 & ZSL & Relation embedding & Text modeling \& GAN & \citet{qin2020generative} \\
\hline
\end{tabular}
\end{table}

With discussion of the LSL methods for RE/C tasks, we categorize the studies of LSL RE/C in Table~\ref{tab4-1-2-1}. While researchers try to utilize prior auxiliary knowledge to alleviate the model's dependence on training data, no matter how sophisticated the knowledge injection strategy is designed, prior knowledge-based models, cannot get rid of two problems: 1. The auxiliary knowledge should be highly relevant to the field of faced relations; 2. This type of model cannot fundamentally solve the need of LSL in rare areas. The RE/C performance will be greatly affected by the quality and relevance of auxiliary knowledge. However, the benefits of prior knowledge are still incorporated into other LSL RE/C methods~\cite{xiong2018one,soares2019matching,zhang2019long,xie2020heterogeneous,qin2020generative}. For example, in semi-supervised methods~\cite{zhang2019long,xie2020heterogeneous}, fresh materials are extracted from the external knowledge library to join the training. However, semi-supervised methods will be trapped in the "cold boot" dilemma without the data-rich source.

The metric-based methods are easy to implement and perfectly adapt to the peculiarity of the long-tail distribution of relation data. This is an excellent choice when the user has a part of sufficient data and needs to learn another piece of relations  but lacks data.  Meta-learner also shows great potential in the RE/C domain. However, RE/C is a multi-factor prediction task (head entity, tail entity, and sentence semantic, etc.), and sentence semantics have a large impact. 
ZSL's RE/C models overcome the scenario without any training data. However, their drawbacks are also obvious: it requires to find a suitable medium task for semantic transfer.


\subsubsection{Graph/Ontology Modeling}
\label{sec4.1.3}

The modeling and calculation of entities (nodes) and relationships (edges) are the main content of structured ontology and knowledge graph modeling. This subsection supplements the previous two subsections from the perspective of graph modeling. We will discuss challenges encountered that include but are not limited to entity and relation modeling.

\begin{figure}[t]
\centering
\includegraphics[scale=0.48]{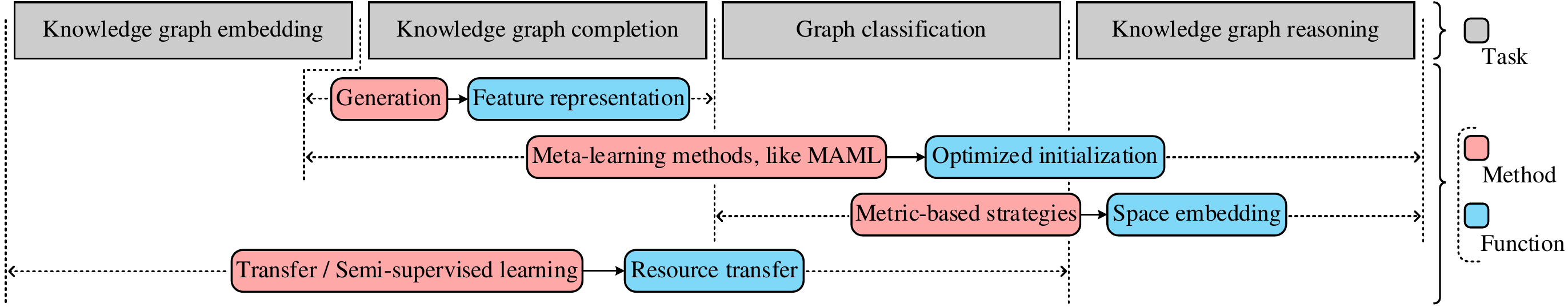}
\caption{Various methods and their corresponding functions, crossing different graph/ontology modeling tasks. The gray blocks refer to graph data modeling tasks, and the red and blue capsules indicate the method type and its primary functions (problems to solve) respectively.}
\label{fig4-1-3-1}
\end{figure}

We mainly introduce graph/ontology modeling researches from four tasks -- Knowledge graph embedding, Knowledge graph completion, Knowledge graph reasoning, and Graph classification. Fig.~\ref{fig4-1-3-1} shows the coverage distribution of different method types on various graph modeling tasks. Most of them are the extension of entity/relation modeling tasks to a more synthesized goal and continue many of the methods introduced in the previous two subsections.

\uppercase\expandafter{\romannumeral1}. \textbf{Knowledge graph embedding} (KGE) aims at learning the low-dimensional representation space for both entities and relations. Compared with the traditional KGE paradigm~\cite{bordes2013translating}, the difficulty of ZSL KGE lies in: Conventional KGE training large-scale seen relation triples \textsc{(head, relation, tail)}, it learns the structured-based representations $\textbf{h}$, $\textbf{r}$, and $\textbf{t}$ for entities $h$, $t$ and relation $r$. However, ZSL KDE must deal with the triples with missed entities, such as: \textsc{(?, relation, tail)}, \textsc{(head, relation, ?)}, even \textsc{(?, relation, ?)}. In description-based ZSL KGE models~\cite{xie2016representation,zhao2017zero}, entity descriptions are leveraged to learn the representations for the novel entities, then supersede the unlearnable structured-based representations. Since the formula of energy function in general KGE model is $E(h, r, t) = \| \textbf{h} + \textbf{r} - \textbf{t} \|$, they need to supplement $\| \textbf{h}_{\textbf{d}} + \textbf{r} - \textbf{t} \|$, $\| \textbf{h} + \textbf{r} -\textbf{t}_{\textbf{d}} \|$, and $\| \textbf{h}_{\textbf{\textbf{d}}} + \textbf{r} - \textbf{t}_{\textbf{d}} \|$ on the basis of $E(h, r, t)$ for cases \textsc{(?, relation, tail)}, \textsc{(head, relation, ?)}, and \textsc{(?, relation, ?)}, where $\textbf{h}_{\textbf{d}}$ and $\textbf{t}_{\textbf{d}}$ are description-based representations for entities $h$, $t$. Although the scope of modeling is extended to "out of KGs", the above methods still need to collect entity description information. For instance, \cite{albooyeh2020out} chooses a semi-supervised strategy to predict the unseen entities based on their relations with the in-sample entities and absorb them into the KG. \cite{zhao2020attention} places these operations as "transition" and "aggregation"  modules.

\uppercase\expandafter{\romannumeral2}. \textbf{Knowledge graph completion/reasoning} (KGC/KGR) in LSL setting means predicting the tail entity $t$ for triple $(h, r, ?)$ with few training samples for relation $r$~\cite{chen2020review}. \cite{zhang2020relation} combines the adversarial and transfer learning strategies to perform precise information delivery from high resource relations to different but related low resource relations. \cite{wang2019tackling} pre-encodes the relation description before conducting meta-learning, and generates extra triples to relieve the problem of data sparsity. The KGC/KGR task can also be regarded as a classification task for relations $r$, and the expected tail entity $t$ is the label of $r$, so that many LSL KGC/KGR models directly follow the common few-shot training paradigm. \cite{lv2019adapting,zhang2020few} applies MAML~\cite{hu2018few} to find well-initialized parameters for multi-hop KGR. \cite{wang2019tackling} utilizes Reptile meta-learner for KGC on the basis of description-based encoder and extra triplet generator. \cite{wang2019meta} modifies MAML method with combining the task-specific information, to enable the model to learn the relationship between different tasks with diverse starting points. The metric-based method is also well applied in FSL KGC/KGR, \cite{du2019cognitive} uses the metric model based on graph neural networks to summarize the underlying relationship of a given instances, then use a  reasoning module to infer the correct answer. In~\cite{xie2019few}, an attention mechanism is added in the graph matching network~\cite{xiong2018one} when describing the interaction between entities and relationships. Additionally, there is research leveraging metric learning methods and meta training algorithms together where the matching network~\cite{vinyals2016matching} is employed to discover similar entity pairs of reference set and the meta-trainer is applied to optimize model parameters~\cite{zhang2019few}.

\begin{table}[t]
\centering
\footnotesize
\caption{Characteristics of LSL Modeling on Graph/Ontology.}
\label{tab4-1-3-1}
\begin{tabular}{p{18pt}|p{18pt}|p{105pt}|p{67pt}|p{40pt}|p{65pt}}
\hline
Task & Type & Technology & Learning Target & Rare object & Reference \\
\hline
\multirow{2}{18pt}[-5pt]{KGE} & \multirow{2}{18pt}[-5pt]{ZSL} & Description embedding & \multirow{2}{67pt}[-5pt]{$E(h, r, t) \leftarrow (h, r, t)$} & \multirow{2}{40pt}[-5pt]{Entity: $h$, $t$} & \citet{xie2016representation,zhao2017zero} \\
\cline{3-3} \cline{6-6}
 & & Semi-supervised learning & & & \citet{albooyeh2020out} \\
\hline
\multirow{2}{18pt}[-10pt]{KGR} & \multirow{5}{18pt}[-22pt]{FSL} & Metric learning & \multirow{6}{67pt}[-25pt]{$t \leftarrow (h, r, ?)$} & \multirow{6}{40pt}[-25pt]{Relation: $r$} & \citet{du2019cognitive,xie2019few} \\
\cline{3-3} \cline{6-6}
 & & Meta-learning & & & \citet{lv2019adapting,wang2019meta,zhang2020few} \\
\cline{1-1} \cline{3-3} \cline{6-6}
\multirow{4}{18pt}[-11pt]{KGC} & & Description embedding \& Meta learning & & & \citet{wang2019tackling} \\
\cline{3-3} \cline{6-6}
 & & Metric learning \& Meta learning & & & \citet{zhang2019few} \\
\cline{3-3} \cline{6-6}
 & & Adversarial learning  \& Knowledge transfer & & & \citet{zhang2020relation} \\
\cline{2-3} \cline{6-6}
 & ZSL & Knowledge transfer & & & \citet{zhao2020attention} \\
\hline
\multirow{4}{18pt}[-1pt]{GC} & \multirow{4}{18pt}[-1pt]{FSL} & Transfer learning & \multirow{4}{67pt}[-1pt]{$y \leftarrow \mathcal{G} = (\mathrm{V}, \mathrm{E}, \textbf{X}) {\rm ^a}$\tnote{a}} & \multirow{4}{40pt}[-1pt]{Graph: $\mathcal{G}$} & \citet{yao2019graph} \\
\cline{3-3} \cline{6-6}
 & & Semi-supervised learning & & & \citet{zhang2018few} \\
\cline{3-3} \cline{6-6}
 & & Metric learning \& Clustering & & & \citet{ma2020few} \\
\cline{3-3} \cline{6-6}
 & & Meta learning & & & \citet{chauhan2019few} \\
\hline
\end{tabular}
\begin{tablenotes}
\footnotesize
\item[a]${\rm ^a}$ $V$ and $E$ refer to the set of the vertices and edges, $\textbf{X}$ is the set of feature vectors of all vertices, $y$ is the label of graph $\mathcal{G}$.
\end{tablenotes}
\end{table}

\uppercase\expandafter{\romannumeral3}. \textbf{Graph classification} (GC) research in the low-shot setting is temporarily rare. However, it covers various methods such as transfer learning~\cite{yao2019graph}, semi-supervised learning~\cite{zhang2018few}, metric learning~\cite{chauhan2019few}, meta-learning~\cite{ma2020few}, etc. In \cite{yao2019graph}, researches incorporate prior knowledge learned from auxiliary graphs to improve classification accuracy on the target graph. An alternative approach utilizes learnable regularization terms to  optimize semi-supervised features while enhancing the generalization of the model when training with few samples \cite{zhang2018few}. \cite{chauhan2019few} applies the metric learning method to project the graph sample on a super-graph prototype space. Additionally, the MAML meta-learner has been modified to capture task-specific knowledge in GC and introduces a reinforcement learning controller to determinate optimal step size \cite{ma2020few}.

To summarize the researches of LSL graph/ontology modeling, Table~\ref{tab4-1-3-1} lists the following information about LSL graph/ontology modeling: related technologies, learning target, and the object which is rare or completely unseen. Graph modeling is a comprehensive task involving a variety of learning objects. When some objects are scarce or completely unseen, many LSL graph modeling methods attempt to exploit other supporting information to make up for the lack of partial training materials~\cite{xie2016representation,zhao2017zero,albooyeh2020out,wang2019tackling,zhang2020relation,zhao2020attention}. When the novel relations or entities have some correlation with the original elements in the KG, the metric-based method~\cite{du2019cognitive,xie2019few,zhang2019few,ma2020few} can learn a reliable embedding space. It can help the model to filter the search range from the open space, which is in line with the target of meta-learning~\cite{lv2019adapting,wang2019meta,wang2019tackling,zhang2019few,chauhan2019few} to find the optimal initial values. It should be noted that the metric-based method requires a stable correlation between the novel sample and the visible sample. At the same time, since natural language forms a natural shared semantic space, metric learning can also make contributions under the setup of ZSL~\cite{albooyeh2020out,zhao2020attention}, which is meta-learner not able to.

\subsubsection{Query Understanding \& Modeling}
\label{sec4.1.4}

The query is the basic operation to access structured data. For LSL query modeling on structured data, the basic strategy consists of actively selecting query and labeled data to create seen classes so that unseen classes can be predicted \cite{xie2016active,xie2017active}. \cite{huang2018natural} extend the training ability of meta-learning framework by effectively creating pseudo-tasks with the help of a relevance function, then solve the semantic parsing problem that maps natural language questions to SQL statements. \cite{elsahar2018zero} leverages triples occurrences in the natural language corpus in an encoder-decoder architecture to build the question generation model for triples of the knowledge base in the "zero-shot" setup, this model utilizes the part-of-speech tags represent a distinctive feature when representing relations in text. \cite{ma2020zero} proposes an approach to zero-shot learning for ad-hoc retrieval models that relies on synthetic query generation, which leverages naturally occurring question/answering (QA) pairs to train a generative model that synthesis queries given a text.

Moreover, LSL methods are also applied in some transformation and parsing task of structured data. \cite{ma2019key} conducts a table-to-text generation task with low-resource training setup. It proposes a novel model to separate the generation into two stages: first, it predicts the key facts from the tables and then generates the text with the key facts. In the second stage, the proposed model is trained with pseudo parallel corpus to alleviate dependence on large-scale training resources. \cite{yu2020few} uses the rule-based and semi-supervised learning strategies to generate weak supervision data for the "few-shot" scenario of conversational query rewriting task. It ten fine-tunes GPT-2 tool\footnote{https://github.com/openai/gpt-2} to rewrite conversational queries.

\subsubsection{Other Scenarios \& Applications}
\label{sec4.1.5}

Structured data modeling is a broad field. In addition to the main application directions described above, LSL has also been adopted in more tasks and scenarios. For example, in the field of bioinformatics, ZSL and FSL methods are used to modeling the topological structure of biomolecules and predict whether a peptide is effective against a given microbial species or not~\cite{gull2020amp0}. \cite{deznabi2020deepkinzero} presents the first ZSL approach to predict the kinase acting on a phosphosite for kinases with no known phosphosite information, which transfers knowledge from kinases with many known target phosphosites to those kinases with no known sites. In addition to bioinformatics domain, \cite{heidari2019holodetect} uses data augmentation function to construct the FSL error detection models that require minimal human involvement.

\subsubsection{Summary}
\label{sec4.1.6}

\begin{figure}[t]
\centering
\includegraphics[scale=0.47]{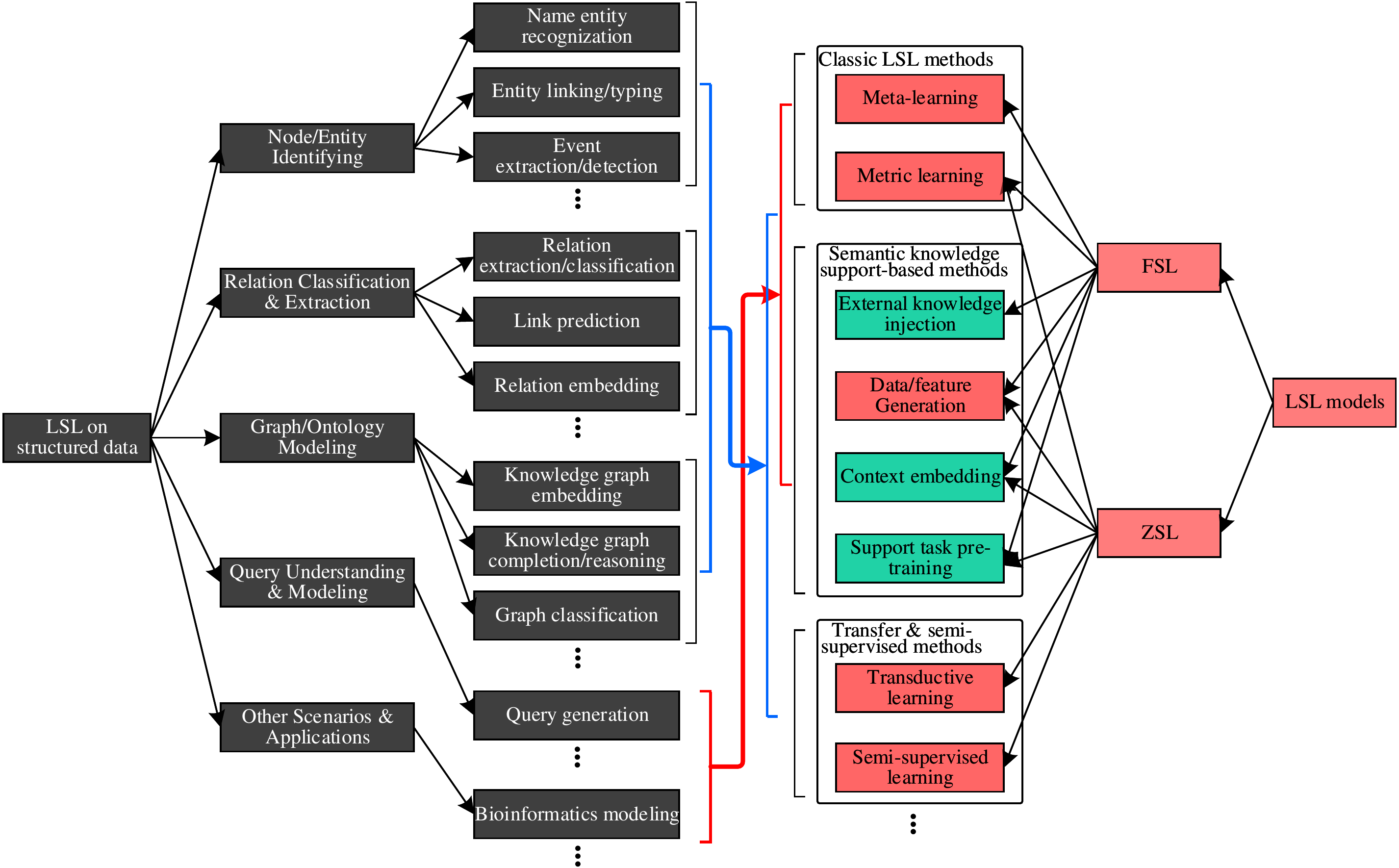}
\caption{A brief taxonomy of LSL for structured data modeling, which expands from two directions. The black box area from left to right represents tasks of LSL structured data modeling, and the colour box area from right to left represents the LSL methods adopted. Among them, the green boxes indicate and approach within LSL specifically for structured data modeling.}
\label{fig4-1-6-1}
\end{figure}

From a high-level view, we can make the following observations:

1. The models and methods used in structured data LSL are split into three major categories: classic LSL methods, LSL methods based on semantic knowledge support, LSL methods with the support of transfer and semi-supervised strategies. The entities, relations, and entire graphs, and related structured data modeling tasks are covered by all three types of methods above (indicated by the blue range bracket and indicator arrow in Fig.~\ref{fig4-1-6-1}). Query understanding and other applications mainly use classic FSL methods and knowledge support technologies (indicated by the red range bracket and arrow in Fig.~\ref{fig4-1-6-1}).

2. Support from external knowledge and contextual semantic model plays a vital role in helping structured data modeling to ease the dependence on a large number of labeled samples. As marked by the green box in Fig.~\ref{fig4-1-6-1}, the structured data LSL models utilize  strategies like external KG injection, context embedding, and assistant NLP task co-modeling to get low-cost support from data sources such as readily available data or existing knowledge libraries. These resources also strengthen the data/feature generation on structured data.

3. For ZSL, the unified semantic environment of natural language in structured data provides a naturally shared space for the transformation of unseen predictions. This shared space provides a similar function to the shared attribute space in the ordinary ZSL paradigm. For FSL, unsupervised knowledge background can not only help structured data LSL models transfer information from the rich sample domain to the few sample domain, but also provide support for the classic FSL method. For example, providing pre-trained encoding for meta-learner~\cite{hu2019few,chen2019meta,baek2020learning} and embedding environment for metric learning methods~\cite{fritzler2019few,soares2019matching}. Moreover, the contextual semantic background enables the metric-based method to be directly applied to ZSL structured data modeling~\cite{huang2018zero,wu2019open}.

4. In addition to the classic FSL method and the embedded shared space method based on knowledge semantics, structured data LSL leverages transfer and semi-supervised learning methods. In addition to transferring information from the rich data field to the poorly-represented data field, LSL often utilizes multi-stage prediction methods to find transitional search terms or narrow the search scope.

After reviewing the many applications of structured data LSL, we can see that the exploration of LSL method in structured data modeling is still in the development stage, and there is still a lot of research space. For example, only meta-learning and metric learning methods are widely utilized. The prospects of other FSL and ZSL methods in structured data modeling tasks still need further investigation. Also, structured data LSL relies on external knowledge and the support of context encoder pre-training. These auxiliary resources often come from manual construction or annotation that has already been completed, which inevitably brings bias~\cite{madras2019fairness,barbosa2019rehumanized}. Therefore, the structured data LSL model also needs to be wary of some risks. For example, 1. Does the corpus used by the pre-trained language model have sufficient coverage? 2. Do the artificially constructed ontology and artificially annotated data for auxiliary tasks bring bias to the LSL model? And so on.

\subsection{Leveraging Structured Data for LSL}
\label{sec4.2}

Structured data such as knowledge systems and ontology libraries are mostly abstract summaries of human experience. Not only can they be modeled under Low-shot setup, but also they provide vital support for LSL in various domains. Pre-defined structured knowledge plays an important role in the realization of LSL for structured knowledge in resource-limited fields. In this section, we further investigate and introduce the supporting techniques of structured knowledge in more areas of LSL tasks. From the perspective of structured data learning and use strategies, we divide knowledge-supported LSL into three types: information injection, feature re-encoding, and objective function feedback. The following descriptions will be unfolded according to the above three categories and different tasks.

\begin{figure}[t]
\centering
\includegraphics[scale=0.62]{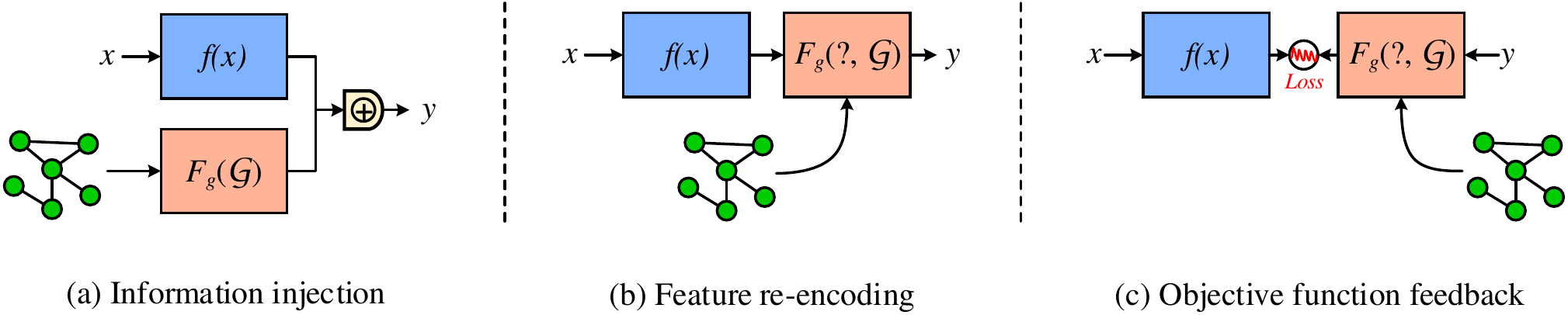}
\caption{Illustration of different knowledge assistant strategies of structured knowledge supported LSL model. (a) The information injection approach which combines the knowledge feature and standard feature; (b) The feature re-encoding approach which re-models the standard feature with knowledge framework; (c) The objective function feedback approach which transform the learning target with knowledge encoding.}
\label{fig4-2-1}
\end{figure}

\uppercase\expandafter{\romannumeral1}. For the learning strategy of information injection, structured ontology knowledge is usually encoded in parallel with conventional standard features, and then combined with standard features  as an information source that supports the final prediction. It can be roughly formulated as:
\begin{equation}
\label{eq4-2-1}
y \leftarrow F_{out} \left\langle F_{g}(\mathcal{G}) \oplus f(x) \right\rangle,
\end{equation}
where operation $\oplus$ denotes some type of combination method, $f(x)$ refers to the standard feature and $F_{g}(G)$ is the encode function for knowledge ontology $\mathcal{G}$, $F_{out} \langle \cdot \rangle$ and $y$ are the prediction function and target. The brief illustration of information injection strategy is also shown in Fig.~\ref{fig4-2-1}(a).

The information-injected knowledge-assisted LSL method is mainly dedicated to designing a coupling strategy of knowledge modules and conventional feature modules. \cite{sui2020knowledge,rios2018few} models knowledge independent to conventional text modeling pipeline and combines knowledge features with standard language model features to be used together for metric-based FSL text classification. In \cite{liu2019semantic,peng2019few,chen2020zero}, in order to optimize the consistency of standard features and knowledge features, before integrating knowledge information and standard features, the model utilizes an auxiliary loss function to optimize the correlation between them. \cite{zhao2020knowledge} designs a specific Mirror Mapping Network, which is leveraged to interactively adjust and map visual features and knowledge features into the same semantic space. In \cite{hu2020semantic}, the gating mechanism is applied to multiply the GCN modeled information of the KG to the visual features of each layer.

\uppercase\expandafter{\romannumeral2}. The learning strategy of feature re-encoding is similar to the information injection approaches, in which standard features are enhanced in the feed-forward process. The difference is that the latter keeps the standard features and knowledge features relatively independent, and it does not use the knowledge system to  also model the standard features. The formulation of feature re-encoding strategy can be written as:
\begin{equation}
\label{eq4-2-2}
y \leftarrow F_{out} \left\langle F_{g} ( f(x), \mathcal{G} ) \right\rangle,
\end{equation}
where the standard features $f(x)$ is re-encoded by function $F_{g}(\cdot)$ with knowledge $\mathcal{G}$, and Fig.~\ref{fig4-2-1}(b) shows the basic data flow of feature re-encoding learning style.

Most of the feature re-encoding knowledge-assisted LSL methods choose not to provide knowledge information directly; however, they leverage knowledge modules to provide a secondary modeling environment for standard features.  The knowledge graph structure can be used to construct an encoding framework and use visual features as the input of knowledge graph modeling, then combine or directly utilize the knowledge modeling features to supply to the predictor \cite{lee2018multi,luo2020context,nayak2020zero}. Alternatively, original features can be used as input to learning dynamic knowledge representation, and use learned knowledge features for prediction \cite{bosselut2019dynamic,kim2019edge,banerjee2020self}. \cite{zhang2019tgg} utilizes attention mechanism to build a graph generation model based on visual features and knowledge graphs, and provide relation propagation features for FSL.  \cite{wang2018zero} regresses the visual features under the knowledge graph representation before performing ZSL classification.  \cite{kampffmeyer2019rethinking} replaces the last layer of the general neural network with knowledge features, then freezes the parameters of that layer and optimizes the weights of other layers, thereby the features of the conventional neural network are forced to absorb knowledge. Finally, some studies apply the strategies of inductive optimization, which makes them  similar to the objective function feedback strategy in the next part.

\begin{table}[t]
\centering
\footnotesize
\caption{Characteristics structured knowledge supporting strategies for LSL application.}
\label{tab4-2-1}
\begin{tabular}{p{35pt}|p{80pt}|p{35pt}|p{95pt}|p{90pt}}
\hline
Strategy & Application & LSL type & Knowledge resource & Reference \\
\hline
\multirow{7}{35pt}[-2pt]{Information injection}& \multirow{2}{80pt}[0pt]{Text classification} & FSL & NELL~\cite{carlson2010toward} & \citet{sui2020knowledge} \\
\cline{3-5}
 &  & FSL \& ZSL & Label descriptions & \citet{rios2018few} \\
\cline{2-5}
 & \multirow{4}{80pt}[-1pt]{Image classification} & \multirow{2}{35pt}[0pt]{FSL} & Categories \& Attributes & \citet{zhao2020knowledge} \\
\cline{4-5}
 &  &  & Categories & \citet{peng2019few} \\
\cline{3-5}
 &  & \multirow{2}{35pt}[0pt]{ZSL} & WordNet~\cite{miller1995wordnet} & \citet{liu2019semantic} \\
\cline{4-5}
 &  &  & Attributes & \citet{hu2020semantic} \\
\cline{2-5}
 & Food image classification & ZSL & Food ingredients & \citet{chen2020zero} \\
\hline
\multirow{7}{35pt}[-5pt]{Feature re-encoding} & \multirow{5}{80pt}[-5pt]{Image classification} & \multirow{3}{35pt}[-5pt]{ZSL} & WordNet~\cite{miller1995wordnet} & \citet{lee2018multi,wang2018zero,kampffmeyer2019rethinking} \\
\cline{4-5}
 &  &  & Visual Genome dataset & \citet{luo2020context} \\
\cline{4-5}
 &  &  & Common sense KG~\cite{speer2017conceptnet,zhang2020transomcs} & \citet{nayak2020zero} \\
\cline{3-5}
 &  & \multirow{2}{35pt}[0pt]{FSL} & ConceptNet 5.5~\cite{speer2017conceptnet} & \citet{zhang2019tgg} \\
\cline{4-5}
 &  &  & Meta-tasks information & \citet{kim2019edge} \\
\cline{2-5}
 & Commonsense QA & \multirow{2}{35pt}[0pt]{ZSL} & \multirow{2}{95pt}[0pt]{ATOMIC~\cite{sap2019atomic}} & \citet{bosselut2019dynamic} \\
\cline{2-2} \cline{5-5}
 & QA &  &  & \citet{banerjee2020self} \\
\hline
\multirow{6}{35pt}[0pt]{Objective function feedback} & \multirow{5}{80pt}[0pt]{Image classification} & \multirow{2}{35pt}[0pt]{ZSL} & WordNet~\cite{miller1995wordnet}, Wikipedia & \citet{liu2018combining} \\
\cline{4-5}
 &  &  & WordNet~\cite{miller1995wordnet} & \citet{wei2019residual} \\
\cline{3-5}
 &  & \multirow{3}{35pt}[-1pt]{FSL} & Class hierarchy & \citet{li2019large} \\
\cline{4-5}
 &  &  & Tabular meta-data & \citet{haney2020fine} \\
\cline{4-5}
 &  &  & Categories & \citet{chen2020knowledge} \\
\cline{2-5}
 & NLU & ZSL & ATOMIC~\cite{sap2019atomic}, etc. & \citet{zhu2020prior} \\
\hline
\end{tabular}
\end{table}

\uppercase\expandafter{\romannumeral3}. The objective function feedback approaches are different from the previous two. They utilize knowledge representation as the learning objective so that features must follow the logic of prior knowledge. The formulation of this strategy is as follows:
\begin{equation}
\label{eq4-2-3}
F_{tgt} \left\langle y, \mathcal{G} \right\rangle \leftarrow F_{out} \left\langle f(x) \right\rangle,
\end{equation}
where the target transform function $F_{tgt}\langle \cdot \rangle$ is used to produce the learning objective with the knowledge graph. In these studies, some utilize an ontology directly as the model output~\cite{li2019large}, and some use the similarity~\cite{haney2020fine,chen2020knowledge} or matching degree~\cite{liu2018combining} between the standard feature and the knowledge feature as the learning target. \cite{zhu2020prior} leverages structured knowledge to customize embedding slot labels for natural language understanding (NLU) tasks. \cite{wei2019residual}  takes word embeddings and knowledge graph as inputs and outputs a visual classifier for each category.

Table~\ref{tab4-2-1} lists the basic information of various knowledge supported LSL models. The structured knowledge  utilized is usually obtained from a third-party prior ontology, the relation of categories or attributes, or a pre-defined category ontology (such as WordNet~\cite{miller1995wordnet}). When structured knowledge is added to general LSL tasks through these methods, they show both advantages and disadvantages:

Information injection approaches are straightforward to implement and understand. Structured knowledge can be modeled offline, which facilitates the realization of plug-and-play auxiliary knowledge  modules. However, information injection often has insufficient interaction with standard features, and lacks feedback on standard features.

Feature re-encoding approaches achieve a high degree of integration of knowledge system and feature modeling. The logical connection found structured knowledge is emphasized. However, the feature re-encoding strategy must bear the corresponding risk when the structured knowledge is biased or does not fit well with the standard feature semantics.

Objective function feedback approaches protects the modeling process of standard features but also provides an optimization mechanism that considers structured knowledge. However, it often needs to adapt to the task-specific output format, and it also needs to worry that knowledge-based re-representation will occur an incorrect interpretation of the original label. Additionally, the expanded output target increases the computational complexity of prediction.

\section{Data Sets of Low-shot Learning for Structured Data}
\label{sec5}

This section mainly introduces the relevant data sets, training benchmarks and evaluation methods of LSL structured data modeling. Different to conventional ZSL and FSL tasks in the field of computer vision, there are many datasets utilized to verify LSL structured data modeling. There are dozens of data sets in total from various tasks. From the perspective of construction strategies, they can be divided into 3 categories: 1. The classic structured data sets follow the training and test settings of LSL; 2. The new data sets to conform to the LSL setting, which is transformed from the classic data set; 3. The new data sets constructed from scratch according to the LSL settings.

\subsection{Classic Data Sets for LSL Usage}
\label{sec5.1}

We follow the content organisation in the method introduction part and introduce LSL-related structured data sets briefly with respect to different application tasks.

For the scenario of entity/node modeling:

\textbf{CoNLL-2003 + SciTech News} datasets provide the challenge of Named Entity Recognition (NER).  \textbf{CoNLL-2003}~\cite{tjong2003introduction} is used as the source domain, and  contains 15000/3500/3700 samples for training/validation/testing.  \textbf{SciTech News}~\cite{jia2019cross} is used as the target domain, which contains 2000 sentences. This task is to identify four types of entities: person, location, organisation, and miscellaneous. The training is on all source samples and test on target sentences~\cite{liu2020zero}.

\textbf{OntoNotes} provides two different datasets. \textbf{OntoNotes}~\cite{che2013named} provides coarse a NER dataset.  It contains  55000 samples in English and Chinese, and has 18 types of named entity.  \textbf{OntoNotes$_{\text{fine}}$}~\cite{gillick2014context} is a fine NER dataset for NER and entity typing, and contains 77 documents and 12017 entity mentions with 17704 labels on 3 levels. Similarly, we also have \textbf{FIGER}~\cite{ling2012fine} and \textbf{BBN}~\cite{weischedel2005bbn} two fine-grained entity typing datasets.

\textbf{Citeseer}~\cite{sen2008collective}, \textbf{Cora}~\cite{sen2008collective}, and \textbf{PubMed}~\cite{velivckovic2017graph} are literature citing datasets, where the nodes are documents, and the edges indicate the citation link between them. There are 3327/4732, 2708/5429, and 19717/44318 nodes/edges with 6, 7, and 3 classes respectively. \textbf{Reddit}~\cite{hamilton2017inductive} contains comments scraped from the Reddit community, it contains more than 230000 nodes for 41 labels. These datasets provide various graph learning tasks~\cite{yao2019graph} including node classification~\cite{zhou2018sparc,zhou2019meta}.

For the scenario of relation learning:

\textbf{FB15K-237}~\cite{toutanova2015representing} and \textbf{NELL}~\cite{xiong2017deeppath} are two classic triplet datasets with 310116 and 154213 triplets in them. The \textbf{FB15K-237} dataset is collected with large knowledge database -- Freebase~\cite{bollacker2008freebase}. The \textbf{FB15K-237} and \textbf{NELL} datasets service the link prediction task~\cite{baek2020learning}, they have 237 and 200 relation labels. Of course, as triple data sets, they also have a considerable number of entities.

\textbf{NYT} dataset~\cite{riedel2010modeling} is often used to evaluate the LSL relation extraction task~\cite{demeester2016lifted,obamuyide2017contextual,yuan2017one,zhang2019long}. It has 53 relation types, and in its training set, there are 522611 sentences, 281270 entity pairs, and 18252 relational facts. In the test set, there are 172448 sentences, 96678 entity pairs, and 1950 relational facts. \textbf{NYT} is also suitable for the tasks with other data types, like entity property identification~\cite{imrattanatrai2019identifying}.

\textbf{SemEval-2010}~\cite{hendrickx2009semeval} is a comprehensive task set for natural language processing. "Task-8" provides 8000 training and 2717 test samples for relation classification task~\cite{soares2019matching,obamuyide2019model}. Another relation classification dataset is \textbf{TACRED}~\cite{zhang2017position}, it has 7500/500 samples for training/testing.

In the scenario of graph modeling, many datasets are shared with the tasks on other data types above. Like \textbf{FB15K-237}~\cite{toutanova2015representing} and \textbf{NELL}~\cite{xiong2017deeppath} datasets are used in knowledge graph completion~\cite{zhang2020relation} and reasoning tasks~\cite{wang2019meta}. Besides these datasets, there are some specific graph classification datasets: \textbf{Little-High}~\cite{riesen2008iam}, \textbf{TRIANGLES}~\cite{knyazev2019understanding}, \textbf{Reddit-12K}~\cite{yanardag2015deep}, and \textbf{ENZYMES}~\cite{borgwardt2005protein}, Little-High has 2250 graphs with average 4.67 nodes, 4.50 edges and 15 classes in each graph; TRIANGLES has 45000 graphs with average 20.85 nodes, 35.50 edges, and 10 classes; Reddit-12K has 11929 graphs with average 391.41 nodes, 456.89 edges, and 11 classes; ENZYMES has 600 graphs with average 32.63 nodes, 62.14 edges, and 6 classes.

\begin{table}[t]
\centering
\footnotesize
\caption{LSL usage information of classic structured datasets.}
\label{tab5-1-1}
\begin{tabular}{p{70pt}|p{100pt}|p{30pt}|p{150pt}}
\hline
Dataset name & Application & LSL type & LSL benchmark references \\
\hline
CoNLL-2003~\cite{tjong2003introduction} ( + SciTech News~\cite{jia2019cross}) & Named entity recognition & ZSL & \citet{hofer2018few,liu2020zero} \\
\hline
OntoNotes~\cite{che2013named} & Named entity recognition & FSL & \citet{fritzler2019few} \\
\cline{2-4}
 & Entity typing & ZSL & \citet{ma2016label,zhou2018zero} \\
\hline
OntoNote$_{\text{fine}}$~\cite{gillick2014context} & Entity typing & ZSL & \citet{zhou2018zero} \\
\hline
FIGER~\cite{ling2012fine} & Entity typing & ZSL & \citet{yuan2018otyper,obeidat2019description} \\
\hline
BBN~\cite{weischedel2005bbn} & Entity typing & ZSL & \citet{zhou2018zero,ma2016label,ren2020neural} \\
\hline
Citeseer~\cite{sen2008collective} & Node classification & FSL & \citet{zhou2018sparc,zhou2019meta} \\
\cline{2-4}
 & Graph classification & FSL & \citet{zhang2018few} \\
\hline
Cora~\cite{sen2008collective} & Node classification & FSL & \citet{zhou2018sparc,zhou2019meta} \\
\cline{2-4}
 & Graph classification & FSL & \citet{zhang2018few} \\
\hline
PubMed~\cite{velivckovic2017graph} & Node classification & FSL & \citet{zhou2018sparc} \\
\cline{2-4}
 & Graph classification & FSL & \citet{zhang2018few} \\
\hline
Reddit~\cite{hamilton2017inductive} & Node classification & FSL & \citet{zhou2019meta,yao2019graph} \\
\hline
FB15K-237~\cite{toutanova2015representing} & Link prediction & FSL & \citet{bollacker2008freebase} \\
\cline{2-4}
 & Knowledge graph completion & FSL & \citet{zhang2020relation} \\
\cline{2-4}
 & Knowledge graph reasoning & FSL & \citet{lv2019adapting,wang2019meta} \\
\hline
NELL~\cite{xiong2017deeppath} & Link prediction & FSL & \citet{bollacker2008freebase} \\
\cline{2-4}
 & Knowledge graph completion & FSL & \citet{zhang2019few,zhao2020attention} \\
\cline{2-4}
 & Knowledge graph reasoning & FSL & \citet{lv2019adapting,wang2019meta} \\
\hline
NYT~\cite{riedel2010modeling} & Entity properties identifying & ZSL & \citet{imrattanatrai2019identifying} \\
\cline{2-4}
 & Relation extraction & FSL & \citet{yuan2017one,obamuyide2017contextual,zhang2019long} \\
\cline{3-4}
 & & ZSL & \citet{demeester2016lifted} \\
\hline
SemEval (task-8)~\cite{hendrickx2009semeval} & Relation extraction & FSL & \citet{soares2019matching} \\
\cline{2-4}
 & Relation classification & FSL & \citet{obamuyide2019model} \\
\hline
TACRED~\cite{zhang2017position} & Relation extraction & FSL & \citet{soares2019matching} \\
\cline{2-4}
 & Relation classification & FSL & \citet{obamuyide2019model} \\
\hline
Little-High~\cite{riesen2008iam} & Graph classification & FSL & \citet{ma2020few,chauhan2020few} \\
\hline
TRIANGLES~\cite{knyazev2019understanding} & Graph classification & FSL & \citet{ma2020few,chauhan2020few} \\
\hline
Reddit-12K~\cite{yanardag2015deep} & Graph classification & FSL & \citet{chauhan2020few} \\
\hline
ENZYMES~\cite{borgwardt2005protein} & Graph classification & FSL & \citet{chauhan2020few} \\
\hline
\end{tabular}
\end{table}

  Given one of the datasets above, FSL models usually sample a subset to ensure that each prediction category contains only a few training samples, thereby fulfilling the Few-shot training requirements. For the setting of ZSL, the researcher divides the prediction categories into the data set into two disjoint parts for training and testing and uses materials like text description, pre-training semantic space etc. to support information transduction from seen labels to unseen labels. Table~\ref{tab5-1-1} lists the information of the classic structured data set used for LSL model evaluation, including the benchmark references of related LSL models. Because of their detail differences in the LSL training settings, it is not informative to compare their performance.

Although the above operations are convenient, there are still some problems in applying these classic data sets to LSL evaluation. First, the LSL splitting and sampling strategies for these data sets are not unified, and researchers will modify them.  Second, many classic data sets are collected from some huge databases, such as Freebase~\cite{bollacker2008freebase} or Wikipedia~\cite{vrandevcic2014wikidata}. When applying external knowledge supporting, LSL models need to avoid inadvertently using information that is duplicated in the test set. This will lead to false test results, but is difficult to enforce in data from these large, connected datasets. Similar problems have attracted wariness in the field of computer vision~\cite{xian2018zero}.

\subsection{New LSL Structured Data Sets}
\label{sec5.2}

The datasets specially constructed for LSL structured data modeling provide a unified data quality, splitting, and measurement, providing a standardized way to measure LSL techniques.

\textbf{FewRel}~\cite{han2018fewrel}. It is the first proposed large dataset which is specifically and well designed for the task of few-shot relation classification, it is annotated by employed crowd-workers. This dataset contains 100 relations and 70000 instances (each category of relation has 700 instances) from Wikipedia. From the perspective of characters, \textbf{FewFel} has 124557 unique tokens in total, and the average number of tokens in each sentence is 24.99. The common split strategy of \textbf{FewFel} dataset is: use 64 relations for training, 16 relations for validation, and 20 relations for testing. There is a benchmark website is available for \textbf{FewFel}~\footnote{https://thunlp.github.io/1/fewrel1.html}.

\textbf{FewRel 2.0}~\cite{gao2019fewrel}. There are two long-neglected aspects in previous few-shot research as uncovered by \textbf{FewFel}: (1) How well models can transfer across different domains. (2) Can few-shot models detect instances belonging to none of the given few-shot classes. In order to investigate these two problems, \textbf{FewRel 2.0} was constructed using  PubMed~\footnote{https://www.ncbi.nlm.nih.gov/pubmed/}. It gathers more 25 relations and 100 instances for each relation. Its also uses data from the  \textbf{SemEval-2010} task 8~\cite{hendrickx2009semeval} dataset.

\textbf{NOTA Challenge} In the NOTA challenge, the developers re-formalize few-shot learning based on the $N$-way $K$-shot setting. For the query instance $x$, the correct relation label becomes $r \in \{r_1, r_2,\dots, r_N , \texttt{NOTA} \}$ rather than $r \in \{ r_1, r_2,\dots, r_N \}$. They also use the parameter NOTA rate to describe the proportion of NOTA queries during the whole test phase. For example, $0\%$ NOTA rate means no queries are NOTA and $50\%$ NOTA rate means half of the queries have the label \texttt{NOTA}. The benchmark list of above two new tasks is available for DA~\footnote{https://thunlp.github.io/2/fewrel2\_da.html} and NOTA~\footnote{https://thunlp.github.io/2/fewrel2\_nota.html} respectively.

\begin{table}[t]
\centering
\footnotesize
\caption{Basic information of new established LSL structured datasets.}
\label{tab5-2-1}
\begin{tabular}{p{58pt}|p{70pt}|p{30pt}|p{60pt}|p{122pt}}
\hline
Dataset name & Application & LSL type & Data resource & Data format \\
\hline
FewRel~\cite{han2018fewrel} & Relation classification & FSL & Wikipedia~\cite{vrandevcic2014wikidata} & Sentence \& Relation name \\
\hline
FewRel 2.0~\cite{gao2019fewrel} & Relation classification & FSL & Wikipedia~\cite{vrandevcic2014wikidata}, PubMed, SemEval-2010 task-8~\cite{hendrickx2009semeval} & Sentence \& Relation name \\
\hline
FewEvent~\cite{deng2020meta} & Event detection & FSL & ACE-2005 corpus, TAC-KBP-2017 Event Track Data, Freebase~\cite{bollacker2008freebase}, Wikipedia~\cite{vrandevcic2014wikidata} & Sentence \& Mention \\
\hline
zero-shot EL~\cite{logeswaran2019zero} & Entity linking & ZSL & Wikia & Sentence \& $<$ Mention, Description $>$ \\
\hline
WikiLinks NED Unseen-Mentions~\cite{onoe2020fine} & Entity linking & ZSL & WikilinksNED dataset~\cite{eshel2017named} & Sentence \& $<$ Mention, Categories $>$ \\
\hline
ManyEnt~\cite{eberts2020manyent} & Entity typing & FSL & FewRel~\cite{han2018fewrel} & Sentence \& Mention, Type \\
\hline
NELL-ZS~\cite{qin2020generative} & Relation learning & ZSL & NELL~\cite{xiong2017deeppath} & Sentence \& $<$ Triple, Description $>$ \\
\hline
NELL-One~\cite{xiong2018one} & Relation learning & FSL & NELL~\cite{xiong2017deeppath} & Triple \\
\hline
Wiki-ZS~\cite{qin2020generative} & Relation learning & ZSL & Wikipedia~\cite{vrandevcic2014wikidata} & Sentence \& $<$ Triple, Description $>$ \\
\hline
Wiki-One~\cite{xiong2018one} & Relation learning & FSL & Wikipedia~\cite{vrandevcic2014wikidata} & Triple \\
\hline
UW-RE~\cite{levy2017zero} & Relation classification & ZSL & Wikipedia~\cite{vrandevcic2014wikidata} & Sentence \& $<$ Relation name, Description $>$ \\
\hline
FB20K~\cite{xie2016representation,zhao2017zero} & Graph representation & ZSL & Freebase~\cite{bollacker2008freebase}, FB15K~\cite{toutanova2015representing} & Triple \& Entity description \\
\hline
\end{tabular}
\end{table}

\begin{table}[t]
\newcommand{\tabincell}[2]{\begin{tabular}{@{}#1@{}}#2\end{tabular}}
\centering
\footnotesize
\caption{Scale and benchmark information of new established LSL structured datasets.}
\label{tab5-2-2}
\begin{tabular}{p{75pt}|p{25pt}|p{25pt}|p{70pt}|p{150pt}}
\hline
\multirow{2}{55pt}[0pt]{Dataset name} & \multicolumn{2}{c|}{Data size} & \multirow{2}{40pt}[0pt]{Measure} & \multirow{2}{180pt}[0pt]{Part of benchmarks} \\
\cline{2-3}
 & Samples & Classes & & \\
\hline
FewRel~\cite{han2018fewrel} & 70000 & 100 & Accuracy & \textit{5-Way 1-Shot: $69.2 \%$; 10-Way 1-Shot: $56.4 \%$}~\cite{han2018fewrel} \\
\hline
FewRel 2.0~\cite{gao2019fewrel} & 100000 & 125 & Accuracy & \textit{5-Way 1-Shot: $56.3 \%$; 10-Way 1-Shot: $43.6 \%$}~\cite{gao2019fewrel} \\
\hline
FewEvent~\cite{deng2020meta} & 70852 & 100 & F1, Accuracy & \textit{5-Way 5-Shot: $73.6 \%$ (F1), $73.9 \%$ (Acc); 10-Way 5-Shot: $61.0 \%$ (F1), $62.4 \%$ (Acc)}~\cite{deng2020meta} \\
\hline
zero-shot EL~\cite{logeswaran2019zero} & 68275 & 16 & Accuracy, Recall@64 & \tabincell{l}{\textit{\textbf{1.} Acc: $63.0\%$, Recall: $82.1\%$}~\cite{wu2019zero} \\ \textit{\textbf{2.} Acc: $54.4\%$, Recall: $71.5\%$}~\cite{logeswaran2019zero}}\\
\hline
WikiLinks NED Unseen-Mentions~\cite{onoe2020fine} & 2.22M & -- & Accuracy & \textit{\textbf{1.} Accuracy: $62.2\%$}~\cite{onoe2020fine}; \textit{\textbf{2.} Accuracy: $71.7\%$}~\cite{wu2019zero}\\
\hline
ManyEnt~\cite{eberts2020manyent} & 100805 & 256(53)${^a}$\tnote{a} & Accuracy & Coarse-grain: $81.14\%$ (1-shot), $90.28\%$ (5-shot); Fine-grain: $79.12\%$ (1-shot), $91.88\%$ (5-shot)~\cite{eberts2020manyent} \\
\hline
NELL-ZS~\cite{qin2020generative} & 188392 & 181 & MRR, Hits@10, etc & \textit{MRR: $0.253$, Hits@10: $37.1\%$}~\cite{qin2020generative} \\
\hline
NELL-One~\cite{xiong2018one} & 181109 & 358 & MRR, Hits@10, etc & \tabincell{l}{\textit{\textbf{1.} MRR: $0.185$, Hits@10: $31.3\%$}~\cite{xiong2018one} \\ \textit{\textbf{2.} MRR: $0.250$, Hits@10: $40.1\%$}~\cite{chen2019meta} \\ \textit{\textbf{3.} MRR: $0.256$, Hits@10: $35.3\%$}~\cite{du2019cognitive}} \\
\hline
Wiki-ZS~\cite{qin2020generative} & 724967 & 537 & MRR, Hits@10, etc & \textit{MRR: $0.208$, Hits@10: $29.4\%$}~\cite{qin2020generative} \\
\hline
Wiki-One~\cite{xiong2018one} & 5859240 & 822 & MRR, Hits@10, etc & \tabincell{l}{\textit{\textbf{1.} MRR: $0.200$, Hits@10: $33.6\%$}~\cite{xiong2018one} \\ \textit{\textbf{2.} MRR: $0.314$, Hits@10: $40.4\%$}~\cite{chen2019meta} \\ \textit{\textbf{3.} MRR: $0.288$, Hits@10: $36.6\%$}~\cite{du2019cognitive}} \\
\hline
UW-RE~\cite{levy2017zero} & 32M & 120 & Precision, Recall, F1 & \tabincell{l}{\textit{\textbf{1.} Precision: $45.9\%$, Recall: $37.4\%$, F1: $41.1\%$}~\cite{levy2017zero} \\ \textit{\textbf{2.} F1: $62.3\%$}~\cite{obamuyide2018zero}} \\
\hline
FB20K~\cite{xie2016representation,zhao2017zero} & 500K & 1341 & Hits@10, etc & \textit{\textbf{1.} Entity prediction: $29.5\%$, Relation prediction: $58.2\%$}~\cite{xie2016representation}; \textit{\textbf{2.} Entity prediction: $42.9\%$}~\cite{zhao2017zero} \\
\hline
\end{tabular}
\begin{tablenotes}
\footnotesize
\item[a]${\rm ^a}$ Coarse-grain(Fine-grain).
\end{tablenotes}
\end{table}

\textbf{FewEvent}~\cite{deng2020meta}. The dataset contains 70852 instances for 19 event types graded into 100 event subtypes in total. In this dataset, each event type is annotated with 700 instances on average. For typical FSL settings, \textbf{FewEvent} is split as: 80 event types are used for training, 10 types for validation, and rest 10 event types for testing. The collection of this dataset includes two phases: 1. the developers first scale up and utilize the event types in 2 existing datasets; 2. the developers then import and extend the new events based on the strategy of anautomatically-labeled event data~\cite{chen2017automatically}, from Freebase~\cite{bollacker2008freebase} and Wikipedia~\cite{vrandevcic2014wikidata}, constrained to some specific domains like music, movie, sports, etc.

\textbf{zero-shot EL}~\cite{logeswaran2019zero}. It is a dataset which applies Wikia~\footnote{https://www.wikia.com.} as construction resource. This dataset provides the task of entity linking which aims to associate the entity mentions and entity dictionary. There are 49K, 10K, and 10K samples in the train, validation, test sets respectively (16 classes, 8/4/4 for Train/Val/Test). Its difference from the other regular entity linking datasets is that the entities in the validation and test sets are from completely different domains than the train set. The entity dictionaries cover different domains, and each contains 10K to 100K entities.

\textbf{WikiLinks NED Unseen-Mentions}~\cite{onoe2020fine}. This dataset is split from another original WikilinksNED dataset~\cite{eshel2017named} to a Unseen-Mentions version. The train, validation, and test sets contain 2.2 million, 10K, and 10K samples respectively. Its definition of unseen-mentions is slightly different from conventional zero-shot entity linking: some entities in the test set can be seen in the training set, however, no mention-entity pairs from test set are observed in the training set, and about 25\% of the entities appear in training set.

\textbf{ManyEnt}~\cite{eberts2020manyent}. This dataset distinctively provides two entity typing modes: coarse-grain and fine-grain. For coarse-grain, there are 256 entity types, and 53 entity types for fine-grain, respectively. Further, the builder also gives a benchmark based on the BERT pre-trained model.

\textbf{NELL-ZS}~\cite{qin2020generative} and \textbf{NELL-One}~\cite{xiong2018one}. These two datasets are constructed from the original KG dataset -- NELL~\cite{xiong2017deeppath}, for zero-shot and one-shot setting respectively. In \textbf{NELL-ZS}, there are 65567 unique entities, 188392 relation triples, and 139/10/32 relation categories for training/validation/testing, there is no intersection between each other. In \textbf{NELL-one}, there are 68545 unique entities, 181109 relation triples, and 358 relation categories where 67 categories are used as one-shot tasks.

\textbf{Wiki-ZS}~\cite{qin2020generative} and \textbf{Wiki-One}~\cite{xiong2018one}. They are similar with above \textbf{NELL-ZS} and \textbf{NELL-one} datasets and based on Wikipedia data~\cite{vrandevcic2014wikidata}. In \textbf{Wiki-ZS}, there are 605812 unique entities, 724967 triples, and 469/20/48 relation categories for training/validation/testing. In \textbf{Wiki-one}, there are 4838244 entities, 5859240 triples, and 822 types of relations, 183 relations are used as one-shot tasks.

\textbf{UW-RE}~\cite{levy2017zero}. It is a relation extraction dataset which consists of 120 relations. Each relation has a set of question templates, each template has both positive and negative relation instances, researchers also produce the description for each relation to support the zero-shot setting. It contains 30 million positive and 2 million negative instances and its split strategy is 84/12/24 relation categories for training/validation/testing.

\textbf{FB20K}~\cite{xie2016representation,zhao2017zero}. This dataset is developed from FB15K dataset~\cite{toutanova2015representing} with 5109 new-added entities which are  built from their descriptions. The training set in \textbf{FB20K} has 472860 triples and 1341 relations, and \textbf{FB20K} has 3 types of test data: 1. ($d$ - $e$), the subject entity is a new entity (Out-of-KG) but the object entity is old (In-KG); 2. ($e$ - $d$), the object entity is new but the subject one is not; 3. ($d$ - $d$), both subject and object entity are new.

These customized data sets provide a unified benchmark and standard experimental specifications for LSL on structured data. Moreover, most of them are reorganized from conventional datasets~\cite{onoe2020fine,qin2020generative,xiong2018one,xie2016representation,zhao2017zero}, or the source of their construction can be easily found~\cite{han2018fewrel,gao2019fewrel,deng2020meta,logeswaran2019zero,levy2017zero}, so as to facilitate comparison of LSL methods. Tables~\ref{tab5-2-1} and \ref{tab5-2-2} list information of the new established LSL structured datasets, such as format, size, and benchmarks. In Table~\ref{tab5-2-1}, "Mention" is the labeled entity that appears in the context, "Triple" refers to the pair with the form $< \texttt{Start\_Entity}, \  \textit{Relation}, \  \texttt{End\_Entity} >$.

\subsection{Databases Used to Assist Unstructured LSL}
\label{sec5.3}

In many LSL studies of unstructured data modeling, structured knowledge provides powerful support. In this section, we briefly discuss some examples of the knowledge database used to support unstructured LSL.

\textbf{WordNet}~\cite{miller1995wordnet} is a large lexical database of English which contains various types of vocabulary. In image recognition tasks, the LSL model often utilizes \textbf{WordNet}'s Noun sub-network to obtain and express the relationship between the recognition objects. The category index in the famous image recognition dataset ImageNet~\cite{russakovsky2015imagenet} is organized according to the Noun structure of \textbf{WordNet}. The Noun part of \textbf{WordNet} starts with 9 "unique beginners", and gradually extends from the "hypernymy" words to the "hyponymy" words, like "Camphor Tree" is the hyponymy word of "Tree", and "Tree" is the hyponymy word of "Plant", forming a tree-like semantic relation structure, thus becoming a natural knowledge graph supporting the LSL model.

\textbf{ConceptNet 5.5}~\cite{speer2017conceptnet} is a larger-scale word and phrase-based semantic network than WordNet. It contains 21 million edges, more than 8 million nodes, and it supports 83 languages. Among them, the English objects include about 1.5 million nodes. Similar to WordNet, this vast database was exploited for class-level graph construction~\cite{zhang2019tgg}, which provides an off-the-shelf knowledge graph connecting objects of natural language edges.

\textbf{ATOMIC}~\cite{sap2019atomic} is a reasoning knowledge graph containing more than 870K logical facts. For various abstract objects, like $X$ and $Y$, this database provides 3 categories of psychological, causal and influence logical relations: "\texttt{If-Event-Then-Mental-State}", "\texttt{If-Event-Then-Event}", and "\texttt{If-Event-Then-Persona}", then subdivided into 9 relation types. The logical facts it provides are often utilized in tasks such as QA, natural language understanding, and to support the training of large-scale natural language pre-training models, like GPT.

\begin{wrapfigure}{l}{0.5\textwidth} \vspace{-5pt}
\centering
\includegraphics[width=0.5\textwidth]{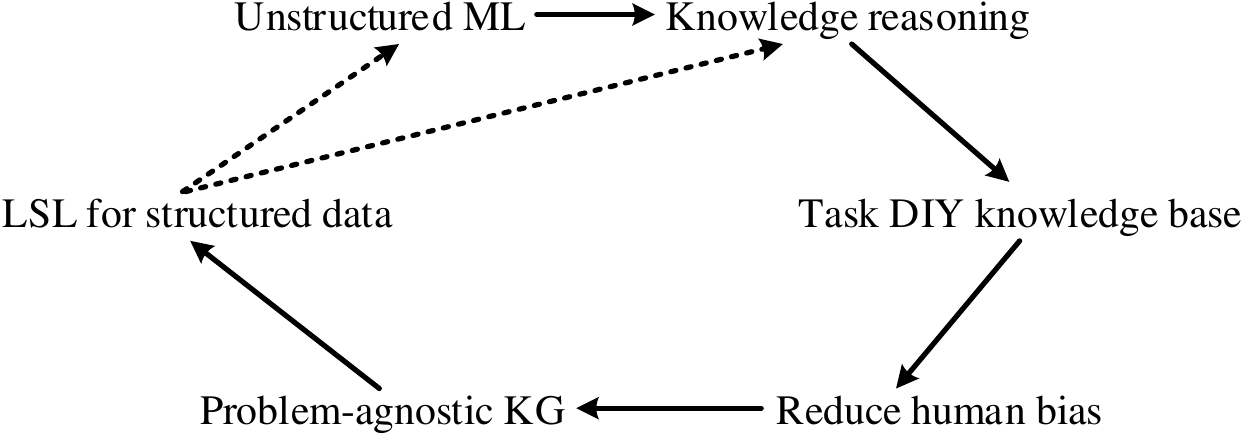}
\caption{The motivation logically closed loop of structured knowledge LSL in typical machine learning tasks.} \vspace{-5pt}
\label{fig5-3-1}
\end{wrapfigure}

Knowledge graphs have already produce an vital cognitive embedding effect in unstructured data ML systems such as computer vision~\cite{gouidis2019review}. Compared to the above-mentioned large-scale knowledge base, many researchers prefer to customize specialized small-scale structured knowledge, because this can serve specific tasks in a targeted manner. Smaller structured knowledge can also save the cost of loading and encoding. At the same time, to minimize the bias caused by manual annotation to the model in the particularly designed knowledge base, it is necessary to design an independent process to establish and represent a "problem-agnostic" KG database. So far, it forms a reasonably closed loop with the previous content, as shown in Fig~\ref{fig5-3-1}.

\section{Relevant Technologies to Low-shot Learning}
\label{sec6}

In this section, we briefly introduce some technologies related to LSL and its application in structured data modeling, and discuss cross-technology combinations and future research trends.

\uppercase\expandafter{\romannumeral1}. \textbf{Transfer learning.} As a long-developed machine learning genre, transfer learning~\cite{pan2009survey} is well-known, and it has many similarities with Low-shot learning. First of all, they are committed to solving the challenge of lack of annotated data in some situations. Second, they all try to help the model to make up for the lack of training data/labels from some external information sources. However, transfer learning has a clear definition of the source and target domains and aims to learn the connection between them  to fine-turn the existing model, and LSL is training a model from scratch to recognize unseen or few data categories. LSL studies often draw on the idea of transfer learning, such as transductive ZSL~\cite{fu2015transductive,song2018transductive,xu2017transductive}, which opens access to some unseen categories to realize feature transfer in semantic space. In the modeling of structured data, "knowledge transfer" techniques~\cite{qu2016named,papanikolaou2019deep,zhao2020attention,yao2019graph} also give sufficient support for LSL.

\uppercase\expandafter{\romannumeral1}. \textbf{Multi-task learning.} Multi-task learning achieves the effect of collaborative optimization by establishing communication between multiple different but related tasks~\cite{ruder2017overview,zhang2018overview}. In LSL, there is research on helping low-shot task training through conventional ML tasks. For structured data, the semantic association between data will be stronger, and the coupling between different knowledge modeling tasks will be higher. It is more conducive to the play of multi-task learning. As far as we know, there is still a lack of related work to joint training on multiple LSL tasks, which is a direction worth paying attention to in the future.

\uppercase\expandafter{\romannumeral3}. \textbf{Lifelong learning.} Lifelong learning embodies the researcher's exploration of models with the ability to learn and acquire memory like human beings~\cite{silver2013lifelong,chen2018lifelong}. Since the modeling of knowledge is generally regarded as an accumulation process, lifelong learning is meaningful for the development of structured data modeling. The meta-learning technology in LSL has an in-depth connection with lifelong learning. Its abstraction of learning tasks helps the model represent and store memory modules for different knowledge modeling tasks.

\uppercase\expandafter{\romannumeral4}. \textbf{Federated learning.} Federated learning is a branch of ML with high practical value. This novel ML setting improves the security of users' data~\cite{yang2019federated}. However, it has always been stuck in performance due to the gap between local data and global data~\cite{li2020federated}. LSL's structured data modeling can become a breakthrough. First, structured data is the primary storage form of users' personal information, and some studies have used meta-learner for model-agnostic sub-task abstraction to reduce the combination loss from local models to the global model~\cite{guha2019one,fallah2020personalized}. Furthermore, the data between different user terminals can be regarded as "unseen" to each other providing an area for  ZSL usage.

\uppercase\expandafter{\romannumeral5}. \textbf{Multi-agent reinforcement learning.} Multi-agent reinforcement learning~\cite{vinyals2019grandmaster,bucsoniu2010multi,foerster2016learning} is dedicated to the optimization of the interaction between multiple intelligent terminals (such as robots~\cite{polydoros2017survey}). The interaction network of multi-agents itself can form a stable structured graph relationship. The combination of LSL technology and multi-agent reinforcement learning may promote the quick learning and cold start capabilities of the agents when there are few trainable interaction records, or the lack of interactive feedback in some special cases.

\section{Industrial Application and Future Challenges}
\label{sec7}

\subsection{Some Industrial Examples of LSL Usage}
\label{sec7.1}

While it is difficult to obtain a wide perspective of usage of low shot learning within industry because of intellectual-property protection, this section contains a few examples that have been documented.

\uppercase\expandafter{\romannumeral1}. \textbf{Personal Account Security.} As a new technology, LSL brings many new possibilities for machine learning. Models that initially required a large scale of data for training now only need a small amount or even zero data to complete training. This leads to new challenges and opportunities in the field of defense and security and has attracted some national attention~\cite{robinson2020few}. FSL technology has been used to detect attackers who steal user credentials~\cite{solano2020few}. Initially biometrics technology based on mouse and keyboard interactionsneeded to collect several hours of interaction records. However, using FSL, it only requires each user to provide 3 to 7 login dialogues. In addition, LSL is also applied to the assessment of personal social privacy status~\cite{li2020graph}. The intervention of FSL dramatically reduces the dependence on labeled data and manual intervention.

\uppercase\expandafter{\romannumeral2}. \textbf{Social Network Analysis.} LSL technology has also been applied to the analysis of large-scale social networks. For instance, the Chinese Tencent Company discovers and analyzes groups on QQ data~\cite{li2019semi}. It is worth noting that large Internet companies have a unique advantage in the field of LSL social network analysis because they have the most comprehensive raw data despite being unlabeled.

\uppercase\expandafter{\romannumeral3}. \textbf{Cyber Security and Defense.} LSL is also utilized to enhance network and computing services defenses against adaptive, malicious, persistent, and tactical offensive threats~\cite{wechsler2015cyberspace}. The scenarios and motivations of these studies are similar to those of ZSL; they have to deal with the situation of unseen attacks and deviations from normal behavior. ZSL technology has also been used to detect unknown cyber intrusions~\cite{zhang2019zero}.

\uppercase\expandafter{\romannumeral4}. \textbf{Recommendation and Intent Recognition.} In addition to the above applications, another industrial-level application of LSL technology is in the field of e-commerce and online services, for user intent identification~\cite{xia2018zero,shen2019zero} and recommendation systems~\cite{li2020few,du2019sequential}. For the former, the original scope of user intention recognition is limited to a small number of predefined intention types. Using the ZSL model, the system can deal with the situation that intents are diversely expressed and novel intents continually be involved~\cite{xia2018zero}.  ZSL technology can also help e-commerce platforms extend user intention recognition to real-time processing scenarios, and  deal with some types of intentions that have never existed during training~\cite{shen2019zero}. For recommendation, FSL technology helps the system make full use of data. It solves some critical contradictions in the potential user recommendation for enterprises: the casual interaction habits of users are difficult to meet the requirement of sufficient interaction record~\cite{li2020few}. In online product and preference recommendation applications, the meta-learning method helps the system overcome the cold-start challenge and enhances the ability to adapt to specific users and scenarios~\cite{du2019sequential}.

\subsection{Identified Challenges}
\label{sec7.2}

Low-shot learning technology and its application on structured data modeling are still based on theoretical research. We briefly describe some of the challenges that need to be considered as LSL on structured data is expanded, as found in various stages of data collection, processing and modelling pipeline.

\uppercase\expandafter{\romannumeral1}. \textbf{Data Collection and Preparation.} It is essential to identify and mitigate bias and ethical issues that may occur in the data collection stage~\cite{barbosa2019rehumanized,olteanu2019social}. Structured data such as databases and knowledge graphs are fundamentally based on human subjective experience,which carries with it bias from collection, sampling and curation. In LSL, due to changes in the learning paradigm, the original data collection and labeling specifications are no longer fully applicable, and due to the reduction in annotation requirements, research needs to be wary of data biases that are built directly into either the limited training samples we have, or into the shared attributes of ZSL. The decrease in training data, incorrect labeling and unnecessary noise will cause  severe damage to the system and possibly exacerbate biases and unfairness in the data.

\uppercase\expandafter{\romannumeral2}. \textbf{Model Training} Many LSL methods use auxiliary knowledge systems to provide help creating teh model~\cite{sui2020knowledge,nayak2020zero,bosselut2019dynamic,banerjee2020self,liu2018combining,wei2019residual,zhu2020prior,zhang2019tgg,lee2018multi,wang2018zero,kampffmeyer2019rethinking}. For structured data modeling, this idea is more generally embodied as "using knowledge to learn knowledge~\cite{rocktaschel2015injecting,demeester2016lifted,obamuyide2017contextual,imrattanatrai2019identifying,zhou2018zero}." One thing we must not ignore is that some prior auxiliary knowledge may also have biases. The selection, representation and embedding of knowledge may all propagate or exacerbate bias.

From a different angle, adversarial attack and defense has absorbed the attention of conventional machine learning in recent years~\cite{dai2018adversarial,yuan2019adversarial,zhang2020adversarial,xu2020adversarial}. Since LSL technology can use fewer resources to complete model training, correspondingly, adversarial attack technology also may utilize fewer materials to confuse LSL models. The models that we are developing may be open to easy attack.

\uppercase\expandafter{\romannumeral3}. \textbf{Prediction and Evaluation.} Some structured LSL data sets provide demos of training and evaluation~\cite{han2018fewrel,gao2019fewrel}. However, the evaluation system of LSL in structured data modeling still needs to be further understood. Under normal circumstances, annotation information of the test data is still necessary for the evaluation stage. In some industrial applications that  face rare situations or invisible categories, new evaluation methods need to be developed.

In addition to the execution and evaluation of a model, it must ultimately be \textit{used} by stakeholders for a particular purpose. The interpretation and transparency problem of black box systems has not yet been completely solved~\cite{ras2018explanation,arrieta2020explainable}. This issue is continued in the structured data modeling of LSL. Thus, when prediction based on a tiny amount of data, the development of the visual interpretation methods for LSL on structured data is a task to be solved in the future. In addition to this, issues surrounding accountability also need to be addressed.

The challenges listed above are broad categories that  LSL on structured data will face for the future. In addition to these, specific problems within LSL will also need to be addressed, such as the problem of "domain shift" in ZSL~\cite{fu2015transductive}, or  the problem of "negative transfer" in FSL~\cite{deleu2018effects}.

\section{Conclusions}
\label{sec8}

In this article, we  focused on the studies of low-shot learning in the field of structured data modeling. We first provided the exact definition of low-shot learning, and then we briefly introduced a variety of general low-shot learning methods. We introduced and compared low-shot learning structured data modeling methods from various application fields such as node/relationship/graph modeling and query. The benefits and assistant strategies of structured data for typical low-shot learning has also been investigated. Then, we introduced the datasets used in structured data low-shot learning. We highlight known "standard" datasets used in this field and newly built ones, respectively. We identified possible areas of additional research such as the combination of low-shot learning and other related machine learning technologies, as well as future work in making LSL commercially viable by analysing industrial applications and challenges.

\section*{acknowledgement}
\label{sec9}

This pure academic work was supported by the Defence Science and Technology Laboratory and the Applied Research Centre at the Alan Turing Institute.

\bibliographystyle{ACM-Reference-Format}
\bibliography{sample-base}


\end{document}